\documentclass[a4paper,authoryear,review]{elsarticle}

\usepackage[utf8]{inputenc}
\usepackage{amssymb,amsmath}
\usepackage{lineno}
\usepackage{url}
\usepackage{hyperref}
\usepackage{caption}
\usepackage{subcaption}
\usepackage{graphicx}
\usepackage{adjustbox}
\usepackage{epstopdf}
\usepackage{gensymb}
\usepackage[table]{xcolor}
\usepackage[normalem]{ulem}
\usepackage{float}
\useunder{\uline}{\ul}{}
\hypersetup{unicode=true,
  colorlinks=true,
  linkcolor=blue,
  citecolor=blue,
  filecolor=red,
  urlcolor=blue,
  breaklinks=true
}

\journal{Journal of Computers and Electronics in Agriculture}

\begin{document}

\begin{frontmatter}

\title{Towards Practical 2D Grapevine Bud Detection with Fully Convolutional Networks}

\author[utn]{Wenceslao Villegas Marset\corref{cor1}}
\ead{diego.villegas@alumnos.frm.utn.edu.ar}

\author[utn]{Diego Sebastián Pérez}
\ead{sebastian.perez@frm.utn.edu.ar}

\author[utn]{Carlos Ariel Díaz}
\ead{carlos.diaz@frm.utn.edu.ar}

\author[utn,conicet]{Facundo Bromberg}
\ead{fbromberg@frm.utn.edu.ar}

\address[utn]{Universidad Tecnológica Nacional. Dpto. de Sistemas de la Información. Grupo de Inteligencia Artificial DHARMa, Mendoza, Argentina.}

\address[conicet]{Consejo Nacional de Investigaciones Científicas y Técnicas (CONICET), Argentina.}

\cortext[cor1]{Corresponding author}

\begin{abstract}
In Viticulture, visual inspection of the plant is a necessary task for measuring relevant variables. In many cases, these visual inspections are susceptible to automation through computer vision methods. Bud detection is one such visual task, central for the measurement of important variables such as: measurement of bud sunlight exposure, autonomous pruning, bud counting, type-of-bud classification, bud geometric characterization, internode length, bud area, and bud development stage, among others. This paper presents a computer method for grapevine bud detection based on a \emph{Fully Convolutional Networks MobileNet} architecture (\textbf{FCN-MN}). To validate its performance, this architecture was compared in the detection task with a strong method for bud detection, Scanning Windows (\textbf{SW}) based on a patch classifier, showing improvements over three aspects of detection: \emph{segmentation, correspondence identification} and \emph{localization}. 
%
The best version of FCN-MN showed a detection F1-measure of $88.6\%$ (for true positives defined as detected components whose intersection-over-union with the true bud is above $0.5$), and false positives that are small and near the true bud. Splits --false positives overlapping the true bud-- showed a mean segmentation precision of $89.3\% (21.7)$, while false alarms --false positives not overlapping the true bud-- showed a mean pixel area of only $8\%$ the area of a true bud, and a distance (between mass centers) of $1.1$ true bud diameters. 
The paper concludes by discussing how these results for FCN-MN would produce sufficiently accurate measurements of bud variables such as \emph{bud number}, \emph{bud area}, and \emph{internode length}, suggesting a good performance in a practical setup.
\end{abstract}

\end{frontmatter}


\section{Introduction}

For decades, viticulturists have been producing models of the most relevant plant processes for determining fruit quality and yield, soil profiling, or vine health, and have been gathering a wealth of information to feed into these models. Better and more efficient measuring procedures have resulted in more information, with its corresponding impact on the quality of model outcomes. 
Such information corresponds to a long list of variables for assessing the state of different parts of the plant, as the one found in the manual published by \cite{awriNDmanual1, awriNDmanual3}.
Most of these variables of interest, however, are still being measured with manual instruments and visual inspection. This results in high labor costs that limit measurement campaigns to only small data samples which, even with the use of statistical inference or spatial interpolation techniques \citep{whelan1996spatial}, restrict the quality of the decisions that agronomists can conduct from them.

Precision viticulture in general \citep{bramley2009lessons}, and computer vision algorithms in particular, have been growing in the last couple of decades mostly due to their potential for mitigating these limitations \citep{seng2018computer, matese2015technology}. These algorithms come along with the promise of an unprecedented boost in the production of vineyard information as well as many expectations not only about possible improvements in the quality of the measurements, but in its potential to produce better models by feeding all this information to big data algorithms.

The present work contributes to this general endeavor with FCN-MN \footnote{Both code and data have been made available online at \url{https://github.com/WencesVillegasMarset/DL4BudDetection}. The shared repository includes both the corpus of images used for training and testing, as well as runnable code for inspecting and visualizing the complete set of results of our experiments, embedding the various models of the FCN-MN detector in variable measurement systems, or re-training the FCN-MN on user provided images.} \citep{long2015fully, shelhamer2017fully}, an algorithm for measuring variables related to one specific plant part: the bud, an organ of major importance as it is the growing point of the fruits, containing all the plant’s productive potential \citep{may2000bud}. 
The present contribution of autonomous bud detection not only enables the autonomous measurement of bud-related variables currently measured by agronomists (see Table~\ref{tab:Table1} for a non-exhaustive list of bud-related variables), but it also has the potential to enable the measurement of novel, yet important, variables that at present cannot be measured manually. One example is the total sunlight captured by buds, which depends on the unfeasible manual task of determining the exact location of buds in 3D space. Although the present work focuses on 2D detection, it could be easily upgraded to 3D by, for instance, integrating 2D detection into the workflow proposed by \citet{diaz2018grapevine}.

Table~\ref{tab:Table1} shows a non-exhaustive list of the main bud-related variables currently measured by vineyard managers \citep{sanchez2005bud, noyce2016basis, collins2020effects}, together with an assessment of the extent to which detection contributes to their measurement. The right-most column (other required operations) indicates the information beyond detection, necessary to complete the measurement, while the middle columns labeled (i), (ii), and (iii) indicate the specific aspects of detection required for that variable: (i) whether it requires a good \emph{segmentation}, i.e., the discrimination of which pixels in the scene correspond to buds and which correspond to non-bud; (ii) a good \emph{correspondence identification}, i.e., discrimination of bud pixels as belonging to different buds; or (iii) a good \emph{localization}, i.e., the localization of the bud within the scene.
For instance, regarding the \emph{bud number} variable, for it to coincide with the detection count, different components detected for the same bud must be bundled together as a single detection. For the \emph{type-of-bud classification}, in addition to correctly identifying components with buds, the segmentation of the part of the image corresponding to the bud must minimize the noise produced by background pixels. Lastly, to measure the \emph{incidence of sunlight on the bud}, localization rather than segmentation is necessary, plus the leaf 3D surface geometry.

\begin{table}[]
 \resizebox{\textwidth}{!}{%
\begin{tabular}{|l|c|c|c|l|}
\hline
\textbf{Variable}               & \textbf{(i)} & \textbf{(ii)} & \textbf{(iii)} & \textbf{Other required operations}\\ \hline
Bud number                     &     & x    &      & none                 \\ \hline
Bud area                       & x      & x    &      & none                 \\ \hline
Type-of-bud classification          & x      & x    &      & plant structure (trunk and canes)     \\ \hline
Bud development stage          & x      & x    &      & classifier over bud mask      \\ \hline
Internode length (by bud detection)       &     & x      & x    & plant structure (trunk and canes)     \\ \hline
Bud volume                    &     &     &      & 3D reconstruction        \\ \hline
Bud development monitoring       & x      & x    & x     & none                   \\ \hline
Incidence of sunlight on the bud       &     & x      & x    & 3D reconstruction, leaves 3D surface geometry \\ \hline
\end{tabular}}
\caption{A non-exhaustive list of important bud-related variables accompanied by an assessment of the extent to which detection contributes to their measurement. The right-most column indicates the information beyond detection necessary to complete the measurement, while the middle columns labeled (i), (ii), and (iii) indicate the three aspects of detection required: segmentation, correspondence identification, or localization, respectively.
}
\label{tab:Table1}
\end{table}

A good detector, therefore, should be evaluated on all three aspects of segmentation, correspondence identification and localization. This is easy for our detector as its implementation first produces a segmentation mask, which is then post-processed to produce correspondence identification and localization. The specific aspects of this approach are detailed in Section~\ref{sec:matmet}. The analysis of detection results presented in Section~\ref{sec:results} shows that this approach is superior to state-of-the-art algorithms for grapevine bud detection. Finally, Section~\ref{sec:discussion} discusses the scope, limitations of the results obtained for bud detection, sufficiency of the performance achieved for the measurement of a selection of variables in Table~\ref{tab:Table3}, as well as the most important conclusions, future work and potential improvements.

\subsection{Related work}
\label{sec:related}
A wide variety of research using computer vision and machine learning algorithms to acquire information about vineyards \citep{seng2018computer} can be found in the literature, such as berry and bunch detection \citep{nuske2011yield}, fruit size and weight estimation \citep{tardaguila2012automatic}, leaf area indices and yield estimation \citep{diago2012grapevine}, plant phenotyping \citep{herzog2014objective, herzog2014initial}, autonomous selective spraying \citep{berenstein2010grape}, and more \citep{tardaguila2012applications, whalley2013applications}. Among the outstanding computer algorithms in recent years, \emph{artificial neural networks} have aroused great interest in the industry as a means to carry out various visual recognition tasks \citep{hirano2006industry, kahng2017cti, tilgner2019multi}. In particular, \emph{Convolutional Neural Networks} (\textbf{CNN}) have become the dominant machine learning approach to visual object recognition \citep{ning2017inception}. Two recent studies have successfully applied visual recognition techniques based on \emph{deep learning networks} to identify viticultural variables to estimate production in vineyards. One of them, \citet{grimm2019adaptable}, uses an FCN to carry out segmentation of grapevine plant organs such as young shoots, pedicels, flowers or grapes. The other, \citet{rudolph2018efficient}, uses images of grapevines under field conditions that are segmented using a CNN to detect inflorescences as regions of interest, and over these regions, the \emph{circle Hough Transform} algorithm is applied to detect flowers.

Several works aim at detecting and locating buds in different types of crops by means of autonomous visual recognition systems. For instance, \citet{tarry2014integrated} presents an integrated system for chrysanthemum bud detection that can be used to automate labour intensive tasks in floriculture greenhouses. More recently, \citet{zhao2018research} presented a computer vision system used to identify the internodes and buds of stalk crops. To the best of our knowledge and research efforts, there are at least four works that specifically address the problem of bud detection in the grapevine by using autonomous visual recognition systems. The research work by \citet{xu2014detection}, \citet{herzog2014initial} and \citet{perez2017image} apply different techniques to perform 2D image detection involving different computer and machine learning algorithms. In addition, \citet{diaz2018grapevine} introduces a workflow to localize buds in 3D space. The most relevant details of each are presented below.

\citet{xu2014detection}’s study presents a bud detection algorithm using indoor captured RGB images and controlled lighting and background conditions specifically to establish a groundwork for an autonomous pruning system in winter. The authors apply a threshold filter to discriminate the background of the plant skeleton, resulting in a binary image. They assume that the shape of buds resembles corners and apply the \emph{Harris corner detector} algorithm over the binary image to detect them. This process obtains a recall of $0.702$, i.e., $70.2\%$ of the buds were detected. 

\citet{herzog2014initial}’s work presents three methods for the detection of buds in very advanced stages of development when the buds have already burst and the first leaves are emerging. All methods are semi-automatic and require human intervention to validate the quality of the results. The best result is obtained using an RGB image with an artificial black background and corresponds to a recall of $94\%$. The authors argue that this recall is enough to solve the problem of phenotyping vines. They also argue that these good results can be explained by the particular green color and the morphology of the already sprouting buds of approximately $2cm$. 

\citet{perez2017image} outlines an approach for the classification of bud images in winter, using \emph{SVM} as a classifier and \emph{Bag of Features} to compute visual descriptors. They report a recall of over $90\%$ and an accuracy of $86\%$ when sorting images containing at least $60\%$ of a bud and a ratio of $20$-$80\%$ of bud vs. non-bud pixels. They argue that this classifier can be used in algorithms for 2D localization of the \emph{sliding windows} type due to its robustness to variation in window size and position. It is precisely this idea that has been reproduced in the present work to implement the baseline competitor to our approach.

Finally, \citet{diaz2018grapevine} introduces a workflow for the localization of buds in 3D space. The workflow consists of five steps. The first one reconstructs a 3D point cloud corresponding to the grapevine structure from several RGB images. The second step applies a 2D detection method using the sliding window and patch classification technique of \citet{perez2017image}. The next step uses a voting scheme to classify each point in the cloud as a bud or non-bud. The fourth step applies the \emph{DBSCAN} clustering algorithm to group points in the cloud that correspond to a bud. Finally, in the fifth step, the localization is performed, obtaining the center of mass coordinates of each 3D point cluster. They report a recall of $45\%$ and a precision of $100\%$ and a localization error of approximately $1.5cm$, or 3 bud diameters. 

Although these research studies represent a great advance in relation to the problem of detecting and localizing buds, they still show at least one of the following limitations: (i) use of artificial background outdoors; (ii) controlled lighting indoors; (iii) need for user interaction; (iv) bud detection in very advanced stages of development; (v) low bud detection/classification recall, and (vi) although some of these works perform some kind of segmentation process as part of the approach, none of them aim to segment the bud or report metrics of the quality of the segmentation performed. These limitations represent a major barrier to the effective development of tools for measuring bud-related variables. 

\section{Materials and Methods}
\label{sec:matmet}


\subsection{Fully Convolutional Network with MobileNet (FCN-MN)}
\label{sec:fcn}

As outlined in the introduction, the approach proposes the use of computer vision algorithms to: (i) \emph{segment} buds by \emph{classifying} which pixels in the scene correspond to buds and which correspond to background (non-buds), (ii) \emph{identify} bud \emph{correspondences} by discriminating those pixels that belong to different buds in the observed scene, and (iii) \emph{localize} each bud in the scene. 

For the segmentation operation, i.e., pixel classification, the fully convolutional network introduced in \citep{long2015fully} is taken as a basis and trained for the specific problem of grapevine bud segmentation. The following section \ref{sec:fcnmn} describes in detail the architecture considered for these networks. The resulting fully convolutional network returns a probability map on the same scale as the original image, where the value of one pixel represents the probability that the corresponding pixel in the input image belongs to a bud. 
To obtain a binary mask, a binarization threshold $\tau$ with values $\{0.1, 0.2, \ldots, 0.9\}$ is applied to each pixel, classifying the pixel as bud (non-bud) if its probability is higher (lower) than $\tau$. To identify bud correspondences, post-processing of this binary mask is performed to determine that two bud pixels correspond to the same bud, as long as they belong to the same connected component, i.e., joined by some sequence of contiguous bud pixels. 
Finally, there are several alternatives for the localization of objects among which are \emph{bounding box}, \emph{pixel-wise segmentation}, \emph{contour} and \emph{center of mass} of the \emph{object} \citep{lampert2008beyond}. In this work the last one was considered, choosing to localize buds by the center of mass of the connected component. 

\subsubsection {Encoder-decoder architecture}
\label{sec:fcnmn}

\begin{figure}
    \centering
    \includegraphics[width=12cm]{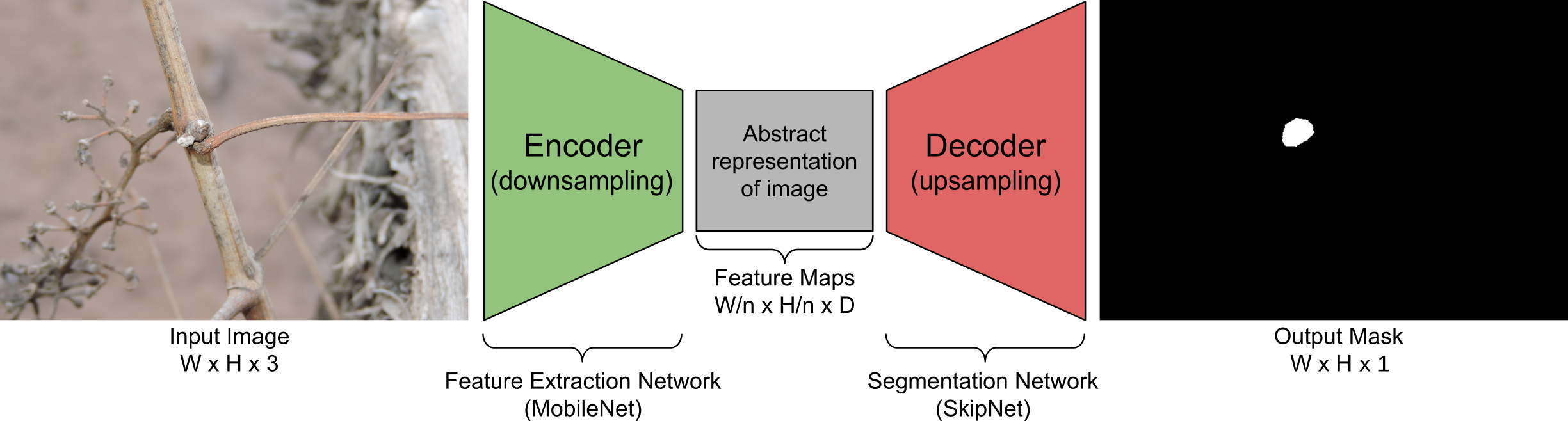}
    \caption{
Diagram of the FCN-MN network architecture proposed in this work, based on the fully convolutional network proposed by \citet{shelhamer2017fully}, replacing its feature extraction encoder with the MobileNet network \citet{howard2017mobilenets}, which produces feature maps with a downsampling factor of $n$. As a decoder for the production of the segmentation map, the SkipNet network \citet{siam2018rtseg} is used, implementing variants 32s, 16s and 8s.
    }
    \label{fig:Figure1}
\end{figure}

For the pixel classifier, the three versions --32s, 16s and 8s-- of the \emph{fully convolutional networks} originally introduced by \citet{long2015fully} were considered, mainly due to their promising results in many image segmentation applications \citep{litjens2017survey, garcia2018survey, kaymak2019brief}. These networks have characteristic architectures with two distinct parts: \emph{encoder} and \emph{decoder} (see Figure~\ref{fig:Figure1}). 

The encoder consists of a convolutional neural network that performs a \emph{downsampling} of an input image into a feature set, by means of convolution operations to produce a set of \emph{feature maps}, i.e., an abstract representation of the image that captures semantic and contextual information, but discards fine-grained spatial information. These operations reduce the spatial dimensions of the image as one goes deeper into the network, resulting in feature maps 1/n the size of the input image, where n is the downsampling factor. The decoder is an \emph{upsampling} subnet, which takes the low-resolution feature map and projects it back into pixel space, increasing the resolution to produce a segmentation mask (or dense pixel classification) with the same dimensions as the input image. This operation is implemented as a network of transposed convolutions with trainable parameters, also known as upsampling convolutions \citep{shelhamer2017fully}.

To refine the segmentation quality, connections that go beyond at least one layer of the network, called \emph{skip connections}, are often used to transfer local spatial information from the internal encoder layers directly to the decoder. In general, these connections improve segmentation results, since they mitigate the loss of spatial information by allowing the decoder to incorporate information from internal feature maps. Their impact may vary depending on the proposed skip architecture. In \citet{long2015fully}, three skip architectures are proposed: 32s without information from internal encoder layers; 16s that adds spatial information from deep encoder layers; and 8s that adds spatial information from deep and less deep encoder layers. The details of these architectures are beyond the scope of this paper, but can be found in \citet{long2015fully} and \citet{shelhamer2017fully}. Since the results reported in the literature are not conclusive regarding which architecture is better, in this work all three alternatives are considered.

In spite of having achieved excellent results in practice, these architectures carry a significant load of computational resources. With this in mind, in this work the VGG encoder of \citet{Simonyan2015VeryDC}, originally proposed by Long for fully convolutional networks, was replaced by the MobileNet network of \citet{howard2017mobilenets},  thus the suffix MN in the name of the FCN-MN algorithm. This network stands out for having only $4.2$ million parameters against the 138 million parameters of VGG, allowing the training and testing process to be considerably faster, with a much lower memory requirement.
This situation was verified by preliminary experimentation, in which indeed MobileNet ended as the fastest, less memory intensive option for our training specification and hardware available. This experimentation is outside the scope of this manuscript and thus further details have been omitted. The use of MobileNet as an encoder in the fully convolutional networks of \citet{long2015fully} is not new, but had already been proposed for the 8s architecture by \citet{siam2018rtseg} in his SkipNet architecture. Technically, \citet{siam2018rtseg}’s proposal is extremely simple; motivating us to extend it to the 16s and 32s architectures originally proposed by \citep{long2015fully}. 

\subsection{Sliding Windows detector}
\label{sec:sw}

\begin{figure}
    \centering
    \includegraphics[width=12cm]{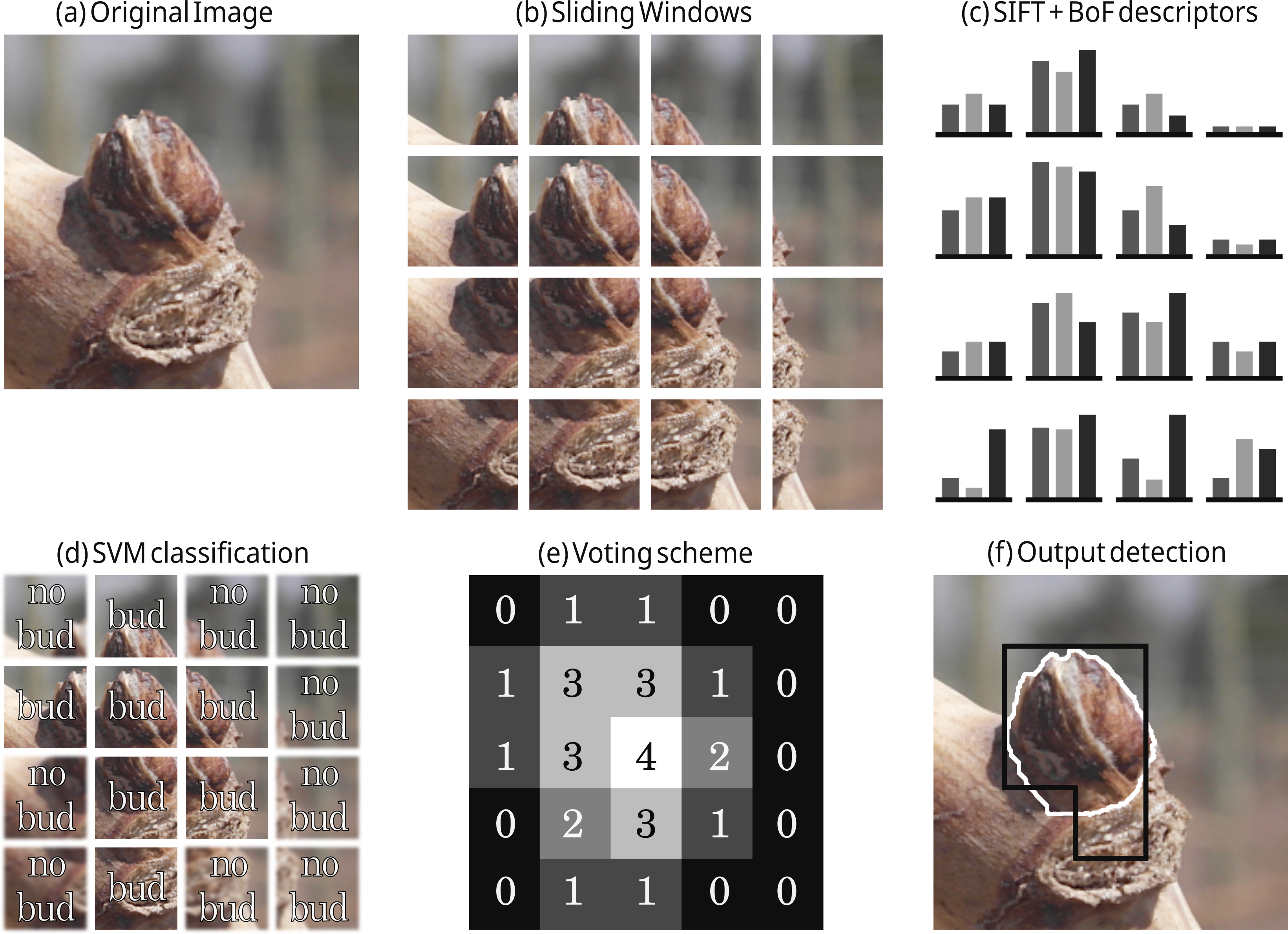}
\caption{
Diagram of the SW bud detection approach based on \citet{perez2017image}. Multiple patches from the input image shown in (a) are extracted via the Sliding Windows algorithm (b). Then, in step (c) a SIFT + BoF descriptor is computed for each patch. These descriptors are classified by an SVM binary classifier (d) in order to determine whether a bud is present on each patch. Finally, the generation of a binary segmentation mask is achieved in two steps. First, in step (e) a voting scheme is applied to each pixel, assigning it one vote for each positive patch it belongs to. Then, the pixel is incorporated in the output segmentation mask if the number of votes it obtained is more or equal to a user given threshold $\nu$. Finally, step (f) shows the output mask corresponding to $\nu=3$ (black) together the ground truth segmentation (white).}
    \label{fig:SW}
\end{figure}

This section describes both, the approach proposed by \citet{perez2017image} for the classification of bud images, and our implementation for detection based on the sliding windows outlined in the original paper, denoted hereon by \textbf{SW}. Details of the six steps of the proposed SW detection procedure are shown in Figure \ref{fig:SW}.

In the present work, different variations of the SW algorithm are contemplated, considering squared windows of the 10 sizes 100, 200, 300, 400, 500, 600, 700, 800, 900 and 1000 pixels, and the four values $\{1, 2, 3, 4\}$ for the voting threshold $\nu$. 
These window sizes were chosen on the basis of the robustness analysis of the classifier presented by \citet{perez2017image} for the window geometry. This analysis shows that the classifier is robust for patches that contain at least $60\%$ of the pixels of a bud, and whose area is composed of at least $20\%$ bud pixels. If we consider extreme cases, i.e., the smallest bud diameter of 100px and the largest of 1600px, window sizes of $100$px and $1000$px could contain at least $60\%$ of the pixels of a bud. In addition, using a $50\%$ displacement, it is guaranteed that at least one patch will contain more than $20\%$ bud pixels, $50$px and $500$px, respectively. The authors argue that a sliding window detection algorithm could easily propose a scheme for choosing window size and displacement to ensure that at some point in the scan the window meets the robustness requirements. 
However, no details are given on how to implement it, so in this paper we only report results for fixed window sizes and $50\%$ vertical and horizontal displacement. As a result, each pixel of the image simultaneously belongs to 4 patches, which justifies the maximum threshold value of $4$. 

For all remaining parameters we considered a single value. The parameter values chosen for the Bag of Features and SVM algorithms (step c) are those of the original publication of \citet{perez2017image}, and are discussed in Section \ref{sec:swtrain} together with details of the algorithms’ training.

\subsection{Model training}
\label{sec:train}

This section provides details of the training process for each approach. In order to contrast both approaches they have been designed to receive the same type of input, i.e., an image of a viticultural scene, and to produce the same outputs, i.e., a binary mask of the same size as the original image whose positive pixels represent bud-type pixels. This allows both algorithms to be trained with the same image collection, which is described in the following section, followed by model-specific training details.

\subsubsection{Image collection}
\label{sec:collection}

The image collection used in this study is the same collection originally used in \citet{perez2017image}, which has been downloaded from \url{
http://dharma.frm.utn.edu.ar/vise/bc} as indicated by the authors. 
The collection corresponds to bud images captured in winter in natural field conditions, on approximately a hundred \emph{Vitis Vinifera} plants from 10 different varieties, all driven by a trellis system. The complete collection consists of 760 images. 
However, in this work, only images containing exactly one bud were kept from the original dataset, resulting in a corpus of 698 images. Cases with more than one bud were scarce for a proper training of the FCN-MN architecture. This may restrict the practical applications of the trained detection model by forcing them to use frames with only one bud. However, training and evaluating a one-bud model lays a strong groundwork for any future work, both academic or technological, that could overcome this limitation by producing the necessary multi-bud training corpus. 
Each image in the corpus is accompanied by the ground truth, that is, a mask of the manual segmentation of the bud. These images and their masks were used during the training and evaluation of the detection models. For this purpose, the image collection was separated into two disjoint subsets: the \emph{train set} with $80\%$ of the images and the \emph{test set} with the remaining $20\%$. This resulted in a train set of 558 images and a test set of 140 images, both with their respective ground truth masks.
\subsubsection{FCN-MN training}
\label{sec:fcntrain}

The 558 images reserved for this purpose were used to train this approach. These images have different resolutions; however, the three proposed FCN-MNs require a fixed size entry. Therefore, all images (including their masks) were scaled to a resolution of $1024 \times 1024$ pixels using a bilinear interpolation method \citep{han2013comparison}. In addition, for the train set images, the pixel RGB intensity values were scaled from [0; 255] to [-1; 1].

Given the small number of images in the train set, two techniques widely used in practice were employed to achieve robust training: \emph{transfer learning} \citep{pan2009survey} and \emph{data augmentation} \citep{shorten2019survey}. The transfer learning process was carried out as follows: (i) the original MobileNet network proposed by \citet{howard2017mobilenets} was implemented; (ii) the network was initialized with the parameters pre-trained on the ImageNet benchmark dataset \citep{kornblith2019better}; (iii) the MobileNet multi-class classification layer was replaced by a binary classification layer; (iv) the network was trained as a bud and non-bud patch classifier in an analogous way to SVM training using the same balanced patch train set used for training SW, after scaling all its images to $224 \times 224$ pixels; and (v) the parameters obtained in the previous step were used to initialize the encoder of our FCN-MN. The data augmentation process was applied on the fly during training, meaning that at each iteration the trainer receives one transformed version of the original image obtained by applying the following seven operations to the original image over parameter values chosen at random with uniform probability: \emph{rotation} of up to $45\degree$; \emph{horizontal shifting} of up to $40\%$; \emph{vertical shifting} of up to $40\%$; \emph{shear} of up to $10\%$; \emph{Zoom} of up to $30\%$; \emph{horizontal flip} and \emph{vertical flip}. Given that there are 200 epochs, the trainer is presented with 200 transformed versions of each image in the corpus, equivalent to one large dataset of 111600 images. 

For the training of the three FCN-MN variants --8s, 16s, and 32s-- it is required to specify the \emph{optimization method} and \emph{dropout} value, two parameters typically defined by the user. In this work, the optimization methods considered were: \emph{Adam} with learning rate $0.001$, $beta1 = 0.9$ and $beta2 = 0.999$; \emph{RMSProp} with learning rate $0.001$ and $\rho = 0.9$; and \emph{Stochastic Gradient Descent} with learning rate $0.0001$ and $momentum = 0.9$. For the dropout case, two values were considered: $0.5$ and $0.001$. These values were pre-selected by preliminary experiments not discussed here.

The best combination of optimization method and dropout was determined in training time over a validation set, using the \emph{4-fold cross validation} approach by 60 epochs and batchsize equal to $4$, varying over the three optimization methods and the two dropout values. The values selected were those that maximize the mean of Jaccard's \emph{Intersection-over-Union} (IoU) \citep{jaccard1912distribution}, a typical assessment measure in segmentation problems. For each combination of optimizer and dropout values the simple mean is reported over $12$ $IoU$s corresponding to the $3$ variants considered in each of the $4$ folds. It can be observed in Table~\ref{tab:Table2} that the combination of parameters with which the highest average $IoU$ is reached is RMSProp with a dropout of $0.001$. Using these parameters, the 8s, 16s, and 32s architectures were trained over 200 epochs and batch size of $4$.

\begin{table}[]
  \centering
       \begin{tabular}{lll}
           \hline
           \multicolumn{1}{|l|}{} & \multicolumn{2}{c|}{\textbf{Mean IoU}} \\ \hline
           \multicolumn{1}{|c|}{\textbf{Optimizer}} & \multicolumn{1}{c|}{\textbf{Dropout = 0.001}} & \multicolumn{1}{c|}{\textbf{Dropout = 0.5}} \\ \hline
           RMSprop & {\ul 0.44253} & 0.3117 \\
           Adam & 0.240277 & 0.315714 \\
           SGD & 0.000886 & 0.00151 \\ \hline
  \end{tabular}
  \caption{
For each combination of optimizer and dropout values the simple mean is reported between $12$ $IoU$s corresponding to the $3$ variants considered in each of the $4$ folds.
}
    \label{tab:Table2}
\end{table}

\subsubsection{SW training}
\label{sec:swtrain}

\begin{figure}
    \centering
    \includegraphics[width=12cm]{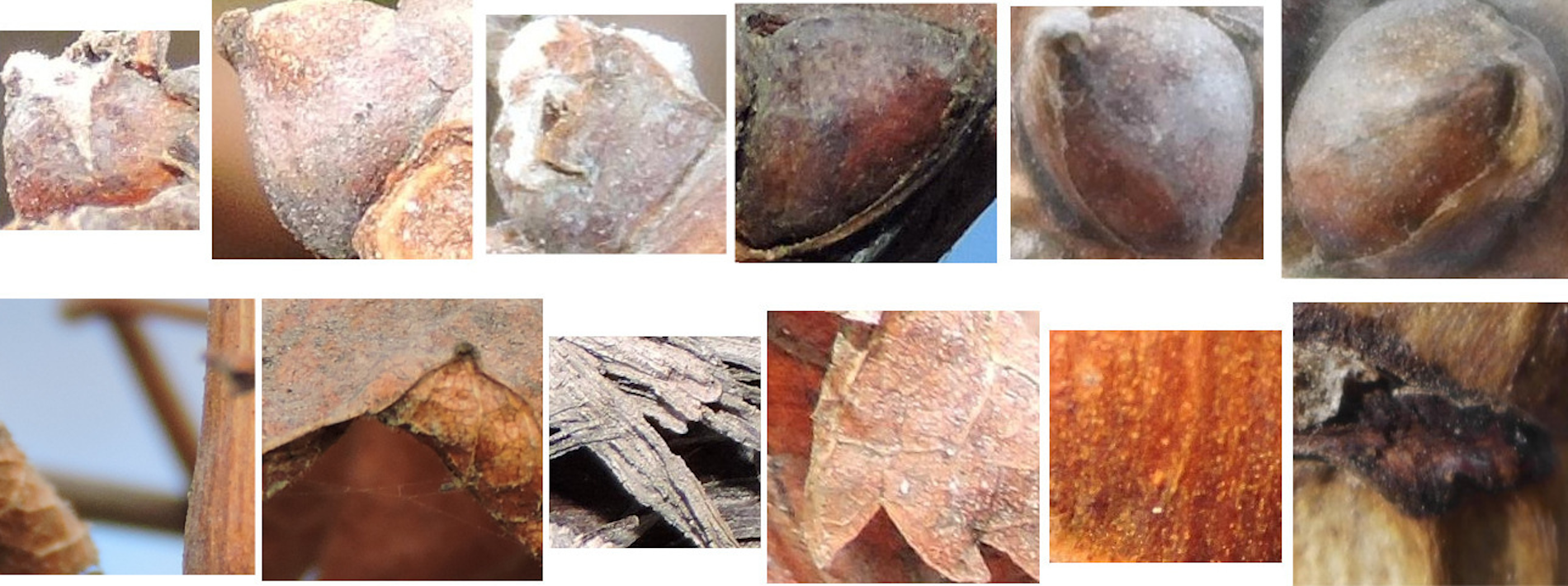}
    \caption{
A sample of the collection of patches used in this work. The first and second rows correspond to bud patches and non-bud patches, respectively. Image extracted from \citet{perez2017image}.
    }
    \label{fig:Figure2}
\end{figure}

The training of SW is conducted in the same way as for the original workflow proposed in \citet{perez2017image}. This involves training a binary classifier to learn the concept of bud versus non-bud from a collection of rectangular patches that may or may not contain a bud. During the training, bud patches must be regions that perfectly circumscribe the bud while non-bud patches must be regions that contain not a single bud pixel (see Figure~\ref{fig:Figure2}). Therefore, to build the patch collection, the $558$ images and their masks were processed following the same protocol as in \citet{perez2017image}, obtaining a total of $558$ patches circumscribing each bud (one per image), and more than $25000$ non-bud patches (the non-bud area is much larger than the area occupied by a bud in the image). The size of these patches is variable, with resolutions between $0.1$ and $2.6$ megapixels for the $100 \times 100$ to $1600 \times 1600$ pixels patches.

From this collection of patches, a balanced patch train set was created, with $558$ patches for each class, where non-bud patches were taken at random from the collection of $25000$ background patches. The training was performed as detailed in the pipeline proposed by \citet{perez2017image}: (i) all SIFT descriptors were extracted from the train set; (ii) BoF was applied with a vocabulary size equal to 25; and (iii) the SVM classifier was trained on the BoF descriptors of each patch using a \emph{Radial Basis Function} kernel, where the value of the $\gamma$ and $C$ parameters was established by means of a 5-fold cross-validation on the same value ranges: $\gamma = \{2^{-14}, 2^{-13}, \ldots, 2^{-7}\}$ and $C = \{2^{5}, 2^{6},\ldots , 2^{14}\}$.

\section{Experimental results} 
\label{sec:results}

In this section we present a systematic evaluation of the quality of our proposed FCN-MN procedure for bud detection over all three aspects of detection required for the measurement of the relevant bud-related variables listed in Table~\ref{tab:Table1}: \emph{segmentation}, \emph{correspondence identification}, and \emph{localization} . 
First, in the following subsection, we present metrics that quantify the quality of these aspects, followed by subsection~\ref{sec:results} that presents the results for the metric values obtained for different experiments over the image test set. 

\subsection{Performance metrics}
\label{sec:metrics}

\subsubsection{Correspondence identification metrics}
\label{subsec:detectmetrics}

Detection of buds is the result of two steps: (i) thresholding of the output masks into a \emph{binary mask}, and (ii) considering each \emph{connected component} of the binary mask as exactly one detected bud. For FCN-MN, the thresholding is done by keeping all pixels of the probabilistic mask with values higher than $\tau$, and for SW this is done through the voting mechanism that keeps all pixels that belong to at least $\nu$ positive patches.
The correspondence identification metrics measure to what extent these detections are \emph{correct} or \emph{incorrect}. Detected components are considered to be correct when most of its mask coincides with the mask of the true bud. This condition is formalized by considering \emph{true positives} as those whose \emph{intersection-over-union} between their masks and the masks of true buds surpasses some threshold $\alpha$. Denoted $IoU$, this coefficient is defined as the area of the intersection between the detected and true masks, normalized by the area of their union. The $IoU$ coefficient  has also appeared in the literature as the Jaccard’s coefficient \citep{jaccard1912distribution}, an alternative to the harmonic mean (a.k.a. F1-measure) of the pixel-wise precision and recall between the detected components and the true mask. The $IoU$ coefficient runs from $0$ when the detection missed completely the true bud, to $1$ when the masks coincide perfectly. Values in between correspond to some true buds missing in the detected mask, or non-buds showing in the detected mask. 
This definition of detection may result in confusion when some detected component correctly detects more than one bud, i.e., some detected component overlaps with an $IoU$ higher than $0.5$ with more than one true bud. This, however, cannot occur in our experiments because the image collection contains only one bud per image. 

This results in the following metrics for correspondence identification, defined for an arbitrary value of the threshold $\alpha$:

\begin{itemize}
\item \textbf{True Positive} ($TP(\alpha)$): number of detected components with $ IoU \geq \alpha $.

\item \textbf{False Positives} ($FP(\alpha)$): number of detected components with $IoU < \alpha$.

\item \textbf{False Negatives} ($FN(\alpha)$):  number of true buds for which there is no true positive detected component, that is, no detected component with $IoU \geq \alpha$. 

\end{itemize}

Rather than reporting these quantities individually, we combine them in the well known precision and recall metrics, denoted as $P_D(\alpha)$ and $R_D(\alpha)$, referred to as \emph{detection-precision} and \emph{detection-recall}, and defined formally as

\begin{align*}
       P_D(\alpha) &= \frac{true\ positives}{true\ positives+false\ positives}
       = \frac{TP(\alpha)}{TP(\alpha)+FP(\alpha)} ,\\
       \\
       R_D(\alpha) &= \frac{true\ positives}{true\ positives+false\ negatives}
       = \frac{TP(\alpha)}{TP(\alpha)+FN(\alpha)} .\\
  \end{align*}

Given these quantities, we also report the \emph{F1-measure}, denoted $F1(\alpha)$, computed as their harmonic average $F1(\alpha) = 2 \times \frac{P_D(\alpha) \times R_D(\alpha)}{P_D(\alpha) + R_D(\alpha)}$.

These correspondence identification metrics provide a strong summarization of the merit of the detection, but may lack some refinements necessary for assessing aspects of the bud detection with impact on the different possible applications of bud related variable measurements. For that, it is paramount to understand the impact of the false positive components. To start then, we distinguish them  between those that overlap a true bud on at least a single pixel, named hereon as \emph{splits}, and those that do not overlap a true bud, named hereon as \emph{false alarms}. Formally,:

\begin{itemize}
\item \textbf{Split} ($S(\alpha)$): number of detected components satisfying $0 < IoU < \alpha$, i.e., are false positives but overlap the true bud. 

\item \textbf{False Alarm} ($FA(\alpha)$): number of detected components with $IoU = 0$, i.e., are false positives and do not overlap a true bud.
\end{itemize}

In the following sections, $\alpha$ is considered to be $0.5$, the most common choice in the literature of detection algorithms, with some minor exceptions considered for a detailed and thorough analysis. To simplify the notation we drop $(\alpha)$ for the cases corresponding to $\alpha=0.5$. For instance, we replace $P_D(0.5)$, $R_D(0.5)$ and $F1(0.5)$ with $P_D$, $R_D$ and $F1$.

\subsubsection{Segmentation metrics}
\label{subsec:segmetrics}

All correspondence identification metrics, including $S$ and $FA$, are based on the rather coarse binary assessment of correct or incorrect. This allows a simple and summarized evaluation but may miss some subtle, pixel-wise errors, and their resulting impact on the measurement of bud related variables. 
A true positive, for instance, could miss that some component may have an $IoU$ much larger than $0.5$, even $1.0$, meaning that its mask is matching perfectly that of the true bud. 
For other non perfect cases, the same $IoU$ can be obtained for many combinations of intersections and unions, with extreme cases of a small detected component completely contained within the true bud presenting the same $IoU$ of a large detected component containing completely the true bud.

A pixel-wise comparison of the masks could also help to assess the quality of false positive detections. For the case of splits, the best case would be one completely enclosed within the true mask, --i.e., presenting not a single false positive pixel--, while covering half minus one of the pixels of the true bud mask. 
For the case of false alarms, a correspondence identification metric may miss how large or small are these false positives. 

The community has proposed several metrics to quantify segmentation errors. The most obvious ones are those that report the \emph{fraction} of the detected mask corresponding to \emph{true positive}, \emph{false positive}, and \emph{false negative} pixels; denoted $TPF$, $FPF$, and $FNF$, respectively. Again, one can simplify the analysis by considering pixel-wise precision and recall, denoted as $P_S$ and $R_S$ and referred to as \emph{segmentation precision}, \emph{segmentation recall}, defined formally as: 

\begin{eqnarray*} 
  P_S &=& TPF / (TPF + FPF),\\
  R_S &=& TPF / (TPF + FNF),
\end{eqnarray*}

which can be combined by their weighted harmonic mean, the well-known \emph{F1-measure} or Dice coefficient. The $IoU$ coefficient is, however, a more natural choice, both for its similarity with the Dice coefficient, and the fact that it was used in the definition of the correspondence identification metrics. 

One could further refine these metrics by applying them, not to the whole mask, but to the individual correspondence identification cases; for instance, by reporting the mean $IoU$ over only true positive components. Also, one could apply them only to splits to assess how bad or good splits are, meaning, how much extra area of the true bud is detected by them. The case of false alarm detections is rather monotonous and not very informative as its precision and recall is always zero. Instead, these components can be better assessed by considering a normalization against bud size to measure their relative size, resulting in the \emph{normalized area}, denoted as $NA$ and defined formally as \emph{the area of the component normalized by the area of the (single) true bud in the image}, with a component’s area corresponding to its total number of pixels.

\subsubsection{Localization metrics}
\label{subsec:locmetrics}

As a complement to the segmentation metrics we consider the localization of the detected components. Mostly useful for false alarms by noticing that false alarms at different distances of the true bud may affect differently the measurement of some bud related variables. In some cases, with proper post-processing (e.g. spatial clustering), the impact of near-by false alarms on the overall error in the measurement of bud related variables may be reduced or could even disappear. 

The selected metric for assessing the localization error of detected components is \emph{normalized distance}, denoted as $ND$ and formally defined as \emph{the distance between the center of mass of the component and the center of mass of the true bud, divided by the diameter of the true bud}, with the bud’s diameter corresponding to the maximum distance between any two bud pixels.

\subsection{Results}
\label{sec:resultados}

This section validates that FCN-MN is a better detector than its SW counterpart through a systematic assessment over each of the metrics defined in the previous section.. 

For a thorough comparison, several cases for each algorithm were considered: training $27$ FCN-MN detectors and $40$ SW detectors over the training set of $558$ images, one for each combination of their respective hyper-parameters. For FCN-MN, these hyper-parameters are the three architectures --8s, 16s, and 32s-- and the $9$ values $\{0.1, 0.2, \ldots, 0.9\}$ for the binarization threshold $\tau$. For SW, in turn, these hyper-parameters are the $10$ patch sizes $\{100, 200, \ldots, 1000\}$ and the $4$ values $\{1, 2, 3, 4\}$ of the voting threshold $\nu$. Once trained, each of these $67$ models were evaluated over the $140$ images reserved for testing purposes, obtaining for each image the detection components.

Table~\ref{tab:Table3} shows the results for the best detectors of each algorithm, reporting all performance metrics of the three aspects of detection over all detected components over the $140$ test images: correspondence identification, segmentation and localization. The first column shows the label of the selected detectors, with the subscript indicating the architecture and patch size for the case of FCN-MN and SW, respectively; and the superscript indicating the thresholds $\tau$ and $\nu$, respectively.

The table includes all metrics defined in Section~\ref{sec:metrics} required for a thorough comparison of FCN-MN against SW. First, four correspondence identification metrics are included: detection-precision $P_D$, detection-recall $R_D$, the F1-measure $F1$, and $S$, the total count split components, all corresponding to an $\alpha$ of $0.5$. Also, seven segmentation metrics are included: the mean and standard deviation (in parenthesis) of the segmentation precision, segmentation recall, and the $IoU$ measure over the $\alpha=0.5$ true positives and splits, denoted in the table by $P_S^{TP}$, $R_S^{TP}$ and $IoU^{TP}$ and $P_S^S$, $R_S^S$ and $IoU^S$ for true positives and splits, respectively; plus the mean and standard deviation of the normalized area for $\alpha=0.5$ false alarms, titled $NA$. 
Finally, the table reports the normalized distance $ND$ of the $\alpha=0.5$ false alarm components, and omits the $ND$ for true positives and splits, assumed too close to the true bud to produce any results of interest. This is confirmed below when their minimum and maximum $NDs$ are reported and discussed. 

The table is a summary, as it includes only a subset of all $27$ FCN-MN cases and all $40$ SW cases. A detector was considered for inclusion in the table if, when compared to its counterparts of the same algorithm, it resulted in the highest value for at least one of the metrics. The corresponding cell was marked in bold in the table. For instance, the detector FCN-MN$_{16s}^{0.6}$ has been included because its detection-precision $P_D$ of $88.6\%$ is the largest among the detection-precision of all $27$ FCN-MN detectors. Similarly, the detector SW$_{700}^3$ has been included because its precision $P_D = 2.5\%$ is the largest among all $40$ SW detectors. 
Also, for all metrics, the best among the FCN-MN detectors (bolded) has been compared to the best among the SW detectors (bolded), and the larger of the two has been underlined.
The table shows an overwhelming improvement of FCN-MN over SW. 
A first analysis of the table shows FCN-MN with larger metrics over SW in all cases, except for the segmentation recall $R_S^{S}$ for splits, for which the SW case has a better (larger) mean of $98.1\%$ compared to the $58.6\%$ for FCN-MN. These improvements are not statistically significant, however, as the large standard deviations of $50.5$ for the FCN-MN cases results in (statistically) overlapping values. 

For the case of correspondence identification metrics $P_D$, $R_D$, $F1$ and $S$, FCN-MN values are overwhelmingly better to those of SW, with the best precisions and recalls of SW all below $6.4\%$ against those of FCN-MN whose values surpass $30.1\%$. For the case of splits one can observe the same pattern, with SW showing the best case of $82$ splits, much larger than the $9$ splits of the best case of FCN-MN.
Although not quite overwhelming, the segmentation metrics of FCN-MN are still larger than those of SW. For instance, for the segmentation precision of true positives $P_S^{TP}$, and split $P_S^{S}$, the FCN-MN over SW improvements are $98.2\%$ versus $94.1\%$, and $98.8\%$ versus $54.2\%$, respectively. 
Finally, for $NA$ and $ND$ (of false alarms), where a smaller value is better, again FCN-MN shows large improvements over SW, with the best values of $NA$ are $0.04$ versus $0.23$, and the best values of $ND$ are $1.10$ versus $5.97$, for FCN-MN versus SW, respectively.

FCN-MN also shows improvements over the mean normalized distances of the true positives and splits. These have been computed but omitted in the table. For FCN-MN the \emph{minimum} and \emph{maximum} mean and standard deviations are $0.038(0.037)$ and $0.055(0.053)$, respectively. Similarly, the FCN-MN minimal and maximal pair for the split components are $0.216(0.138)$ and $0.482(0.212)$, respectively. As predicted, all rather small, with both the minimum and maximum mean distance falling well within one diameter of a true bud, for all cases.
For the SW detectors, the min/max pair of mean normalized distances for the true positive components is $0.045(0.023)$/$0.210(0.076)$, and for splits components is $0.412(0.210))$/$3.250(5.961)$, respectively. 
As can be observed, again FCN-MN shows an improvement over SW, with a minor statistically significant overlap of their min/max intervals for both the true positives and split cases.

 \begin{table}[]
    \scriptsize
    \tiny
    \centering
    \setlength{\tabcolsep}{1.5pt}
    \renewcommand{\arraystretch}{1.08}     
    \begin{adjustbox}{angle=90}
     \resizebox{.76\textheight}{!}{%
      \begin{tabular}{lcccccccccccc}
       \hline
       \multicolumn{1}{|l|}{\textbf{Detector}} & \multicolumn{1}{c|}{\textbf{$P_D$}} & \multicolumn{1}{c|}{\textbf{$R_D$}} & \multicolumn{1}{c|}{\textbf{$F1$}} & \multicolumn{1}{c|}{\textbf{$S$}} & \multicolumn{1}{c|}{\textbf{$P_S^{TP}$}} & \multicolumn{1}{c|}{\textbf{$R_S^{TP}$}} & \multicolumn{1}{c|}{\textbf{$IoU^{TP}$}} & \multicolumn{1}{c|}{\textbf{$P_S^S$}} & \multicolumn{1}{c|}{\textbf{$R_S^S$}} & \multicolumn{1}{c|}{\textbf{$IoU^S$}} & \multicolumn{1}{c|}{\textbf{$NA$}} & \multicolumn{1}{c|}{\textbf{$ND$}} \\ \hline
       \textbf{$FCN_{8s}^{0.1}$} & 39.6 & 92.1 & 55.4 & 17 & 78.8(9.1) & 97.6(6.8) & 76.9(8.9) & 63.0(32.3) & \textbf{58.6(50.5)} & 22.3(21.1) & 0.17(0.81) & 7.61(7.54) \\
       \textbf{$FCN_{8s}^{0.2}$} & 59.4 & {\ul \textbf{95.0}} & 73.1 & 13 & 83.5(9.5) & 96.7(6.2) & 80.8(9.3) & 70.5(36.7) & 36.1(44.7) & 15.2(17.3) & 0.29(0.88) & 4.73(5.45) \\
       \textbf{$FCN_{8s}^{0.3}$} & 63.0 & {\ul \textbf{95.0}} & 75.8 & 15 & 87.4(8.5) & 95.0(7.8) & 83.1(9.1) & 75.2(36.1) & 28.4(42.7) & 11.8(16.9) & 0.28(0.77) & 3.98(4.74) \\
       \textbf{$FCN_{8s}^{0.4}$} & 69.3 & 93.6 & 79.6 & {\ul \textbf{9}} & 90.1(7.9) & 93.8(6.9) & {\ul \textbf{84.7(8.2)}} & 71.1(32.4) & 54.2(38.7) & {\ul \textbf{29.5(17.7)}} & 0.29(0.76) & 3.54(4.47) \\
       \textbf{$FCN_{8s}^{0.9}$} & 70.1 & 82.1 & 75.7 & 34 & {\ul \textbf{98.2(5.1)}} & 75.1(11.3) & 73.8(10.6) & {\ul \textbf{98.8(7.2)}} & 17.7(19.7) & 17.1(18.4) & 0.24(0.5) & 3.80(5.66) \\
       \textbf{$FCN_{16s}^{0.4}$} & 80.6 & 89.3 & 84.7 & 10 & 88.6(8.2) & 93.3(8.5) & 82.8(8.7) & 78.6(33.6) & 32.0(33.5) & 22.0(16.2) & {\ul \textbf{0.04(0.09)}} & 3.80(5.08) \\
       \textbf{$FCN_{16s}^{0.6}$} & {\ul \textbf{88.6}} & 88.6 & {\ul \textbf{88.6}} & 10 & 92.8(6.7) & 89.3(10.2) & 83.1(9.4) & 89.3(21.7) & 26.9(34.1) & 18.6(19.5) & 0.08(0.11) & {\ul \textbf{1.10(0.65)}} \\
       \textbf{$FCN_{32s}^{0.1}$} & 30.1 & 88.6 & 44.9 & 32 & 71.5(10.1) & {\ul \textbf{98.2(5.5)}} & 70.2(9.1) & 69.1(30.2) & 46.1(48.1) & 19.2(19.5) & 0.14(0.66) & 4.62(5.59) \\ \hline
       \textbf{$SW_{200}^{4.0}$} & 2.1 & \textbf{6.4} & 3.2 & 142 & 72.2(6.0) & 75.1(8.3) & 58.1(6.4) & 44.5(31.9) & 40.1(33.8) & 17.6(14.0) & 1.00(1.78) & 8.68(6.58) \\
       \textbf{$SW_{100}^{4.0}$} & 0.3 & 1.4 & 0.5 & 196 & 59.5(4.6) & 85.5(16.7) & 53.4(2.9) & \textbf{54.2(34.8)} & 17.1(22.3) & 11.0(12.4) & \textbf{0.23(0.59)} & \textbf{5.97(6.51)} \\
       \textbf{$SW_{1000}^{4.0}$} & 1.4 & 2.1 & 1.7 & \textbf{82} & 65.6(5.1) & 74.0(6.0) & 53.1(3.1) & 20.4(17.3) & 67.0(32.1) & 16.3(12.4) & 13.87(21.8) & 7.15(5.2) \\
       \textbf{$SW_{700}^{3.0}$} & \textbf{2.5} & 5.0 & \textbf{3.4} & 109 & 64.0(6.0) & 85.1(7.7) & 57.2(4.8) & 15.8(14.0) & 82.1(26.1) & 13.6(9.1) & 15.95(28.85) & 8.10(4.79) \\
       \textbf{$SW_{600}^{2.0}$} & 0.3 & 0.7 & 0.4 & 135 & 54.3(--) & \textbf{97.1(--)} & 53.4(--) & 10.2(10.0) & 91.6(21.0) & 9.8(9.5) & 20.63(38.89) & 7.94(4.39) \\
       \textbf{$SW_{500}^{1.0}$} & 0.0 & 0.0 & 0.0 & 140 & 0.0(--) & 0.0(--) & 0.0(--) & 8.4(9.6) & {\ul \textbf{98.1(9.6)}} & 8.3(9.5) & 17.39(30.06) & 7.22(4.04) \\
       \textbf{$SW_{500}^{4.0}$} & 0.4 & 0.7 & 0.5 & 119 & \textbf{94.1(--)} & 70.1(--) & \textbf{67.2(--)} & 27.9(22.3) & 60.2(31.1) & \textbf{19.7(12.0)} & 5.90(8.43) & 9.53(5.76) \\ \hline
      \end{tabular}
     }
    \end{adjustbox}
     \caption{
    Correspondence identification, segmentation and localization metrics for the best FCN-MN and SW detection models. Each column shows bolded cells corresponding to the cell with the best metric among all FCN-MN rows and the cell with the best metric among SW rows, and underlined cells corresponding to the best among all combined models, i.e., the best of the column. Columns $P_D$, $R_D$, $F1$ and $S$ show results for the \emph{Correspondence identification metrics} detection-precision, detection-recall, F1-measure and number of images with splits, respectively: Columns $P_S^{TP}$, $R_S^{TP}$ and $IoU^{TP}$ (resp. $P_S^S$, $R_S^S$ and $IoU^S$) correspond to the \emph{segmentation metrics} mean segmentation precision, mean segmentation recall, and mean $IoU$ measure over all true positive components (resp. split components), with standard deviations in parenthesis (undefined cases denoted by ``-"); and Columns $NA$ and $ND$ show the mean $NA$ and mean $ND$ over all false alarm components. 
 }
 \label{tab:Table3}
\end{table}

\subsubsection{Detailed analysis of correspondence identification metrics}
\label{sub:compFCNSW}

\begin{figure}

    \centering
  \begin{subfigure}[b]{0.97\textwidth}
       \centering
       \includegraphics[width=\textwidth]{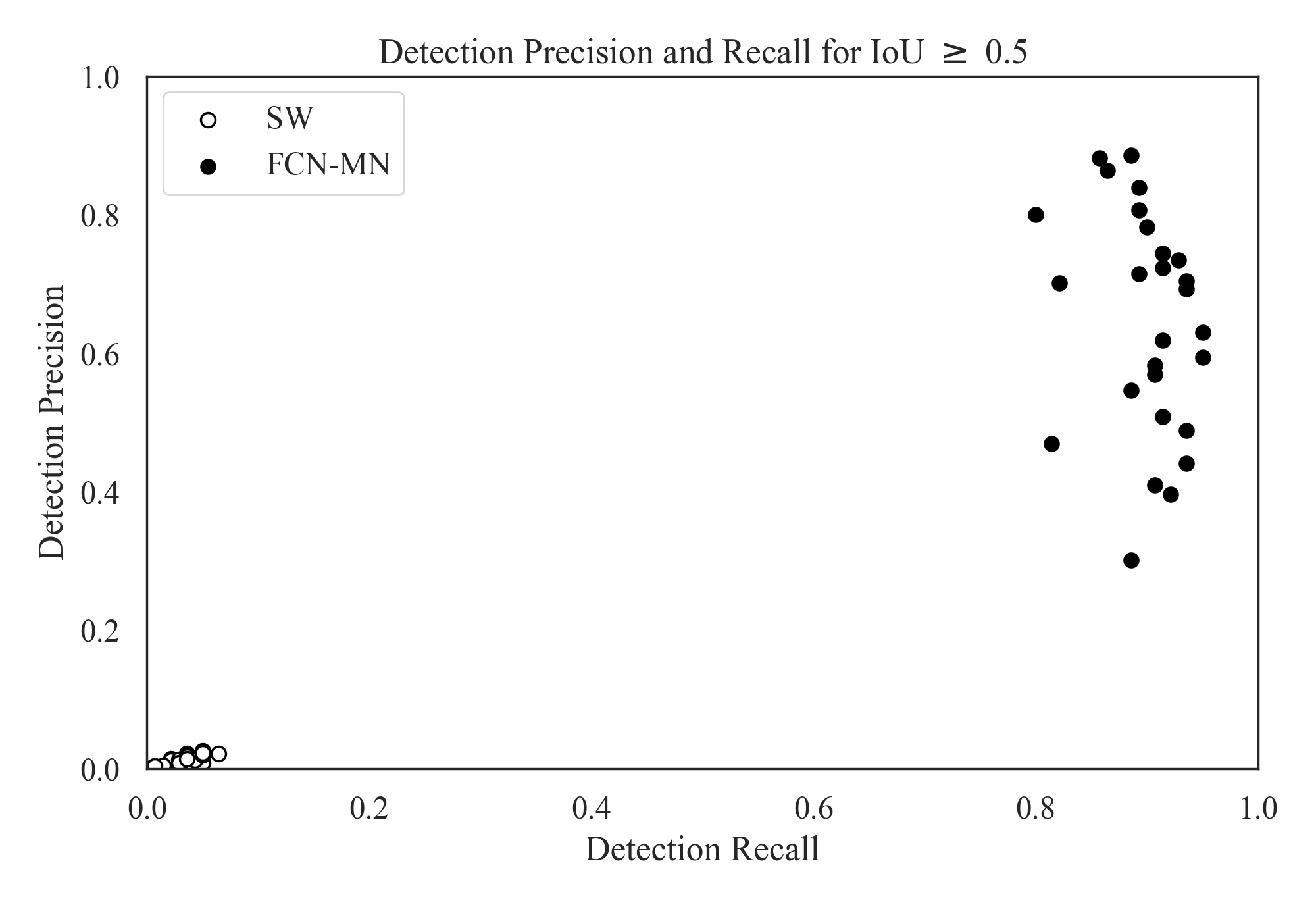}
       \caption{}
       \label{fig:Figure3-a}
  \end{subfigure}
  \hfill
  \begin{subfigure}[b]{0.97\textwidth}
       \centering
       \includegraphics[width=\textwidth]{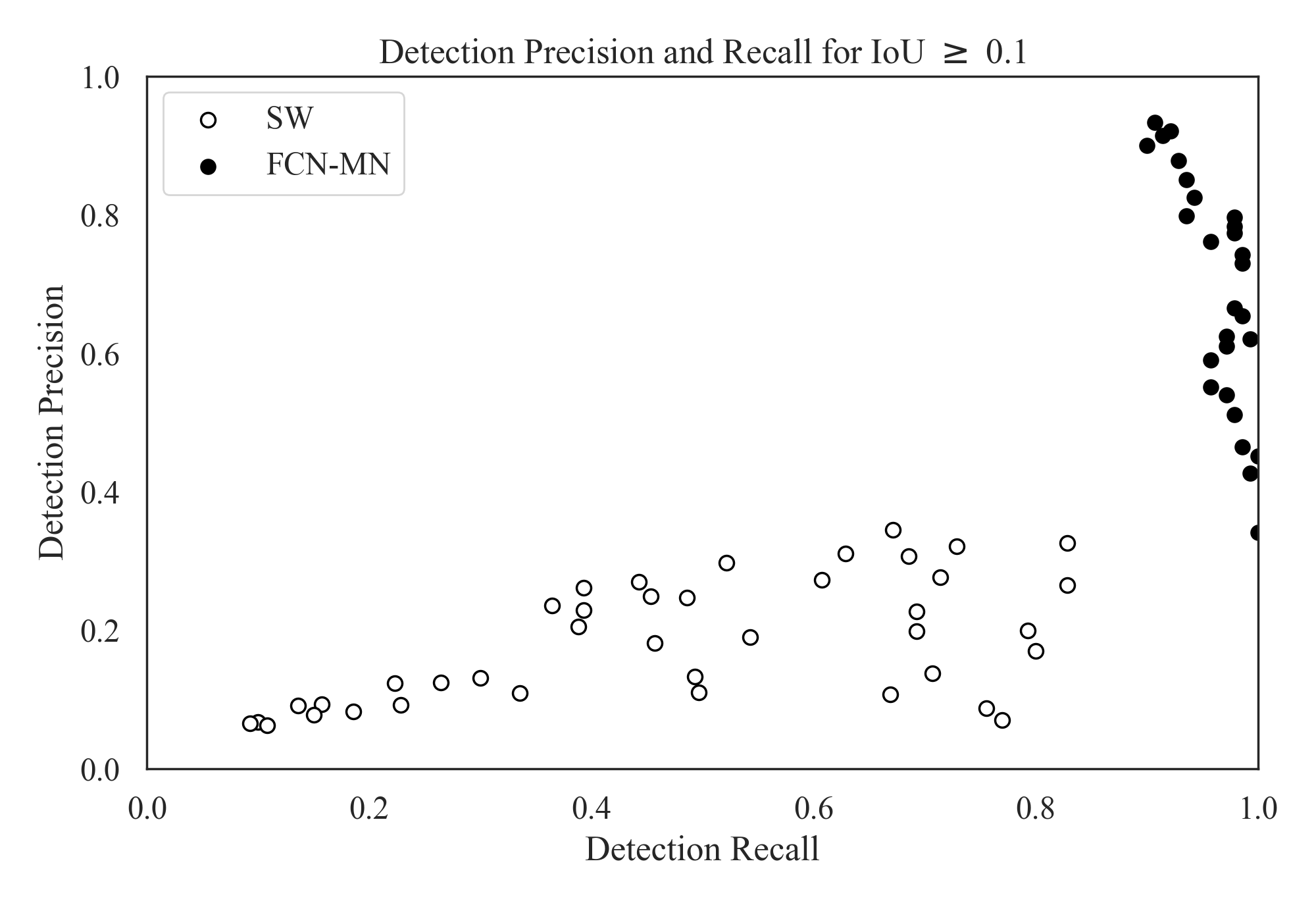}
       \caption{}
       \label{fig:Figure3-b}
  \end{subfigure}
    \caption{
Scatterplots of detection-precision $P_D(\alpha)$ versus detection-recall $R_D(\alpha)$, with  results for $\alpha=0.5$ and $\alpha=0.1$ shown in Figure (a) and (b), respectively. Results for FCN-MN and SW are shown in black and white dots, respectively. Each dot represents the detection-precision $P_D$ and detection-recall $R_D$ for some particular configurations of hyper-parameters among all models ($27$ for FCN-MN and $40$ for SW).
    }
    \label{fig:Figure3}
\end{figure}

Graphically, one could expect a better combined analysis of detection-precision and detection-recall than could be obtained by comparing the F1-measure. This is shown as a scatter plot in Figure~\ref{fig:Figure3-a}, a graphical representation of a non-summarized version of the second and third columns of Table~\ref{tab:Table3}. Each dot in the plot is located according to the detection-precision and detection-recall, and the color black or white, whether it corresponds to an FCN-MN or an SW detection model.

The graph reinforces the clear and undisputed improvements of FCN-MN over SW already shown in the table, with overwhelmingly larger detection precisions and recalls.

One concern that may arise is the confidence one can ascribe to such overwhelmingly bad results of the SW correspondence identification metrics. One possible explanation may arise by noticing the large number of splits of SW. Splits are components that could not pass the $IoU \geq 0.5$ condition but are overlapping the true bud. This suggests that SW may be producing too many small detections of the bud, all of which could not make the cut for $\alpha=0.5$. This can be confirmed by observing in Table~\ref{tab:Table3} the mean $IoU$ of splits for the SW models, all of which are well below $50\%$, with the maximum at $19.7\%$. We complement this by also considering correspondence identification metrics with a smaller $\alpha$ of $0.1$, whose precision-recall scatterplot is  shown in Figure \ref{fig:Figure3-b}. With similar results for  FCN-MN, the graph shows clear improvements for $SW$, with precisions reaching almost $40\%$ and recalls above $80\%$. 
The increase in the detection-precision proves that many of the $\alpha=0.5$ splits are true positives for $\alpha=0.1$. This may even result in some true buds with not a single true positive for the case of $\alpha=0.5$, may now have one, which explains  the increase in detection-recall.

\subsubsection{Detailed analysis of segmentation metrics}

\begin{figure}%
  \centering
  \begin{subfigure}[b]{0.97\textwidth}
       \centering
       \includegraphics[width=\textwidth]{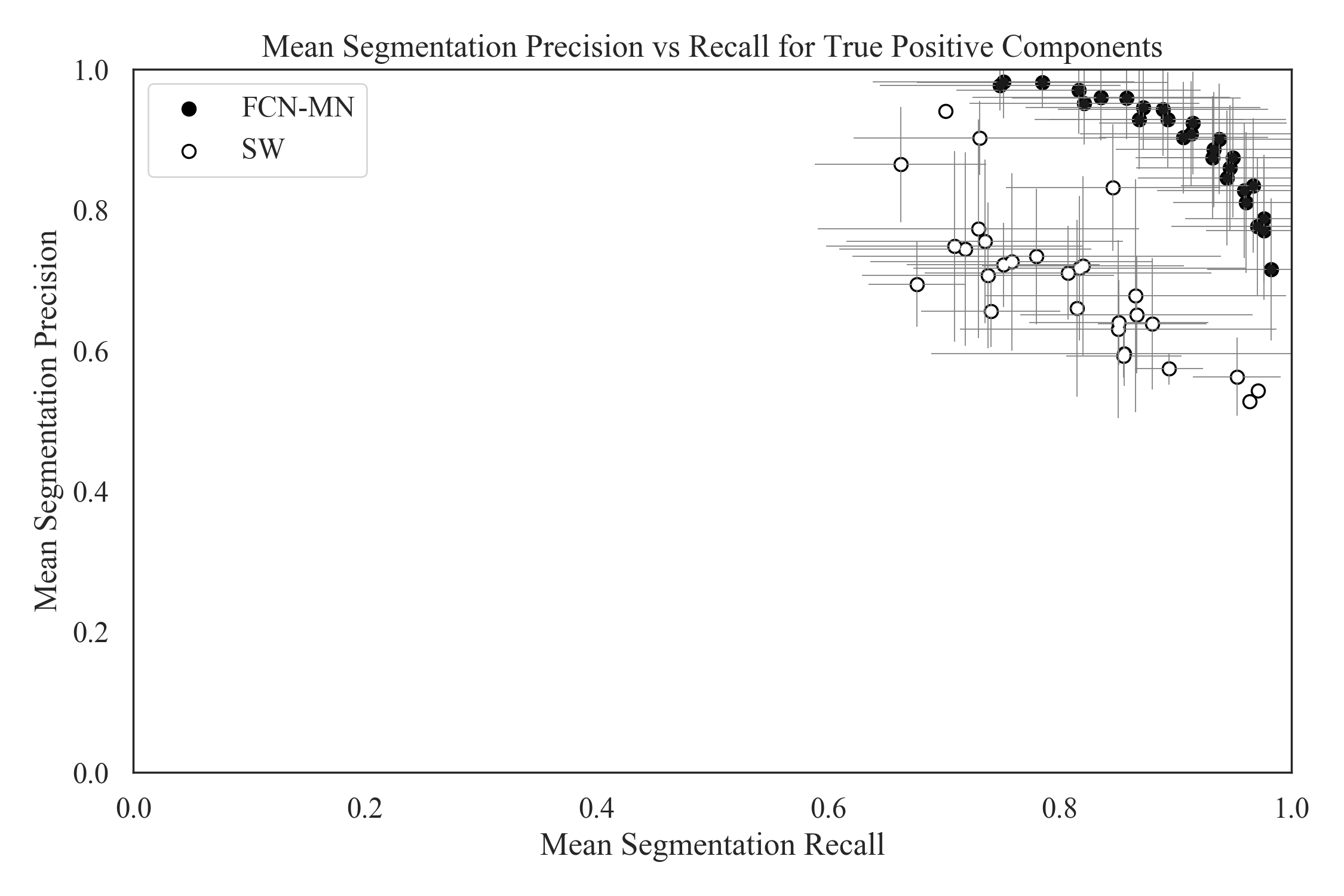}
       \caption{}
       \label{fig:Figure5-a}
  \end{subfigure}
  \hfill
  \begin{subfigure}[b]{0.97\textwidth}
       \centering
       \includegraphics[width=\textwidth]{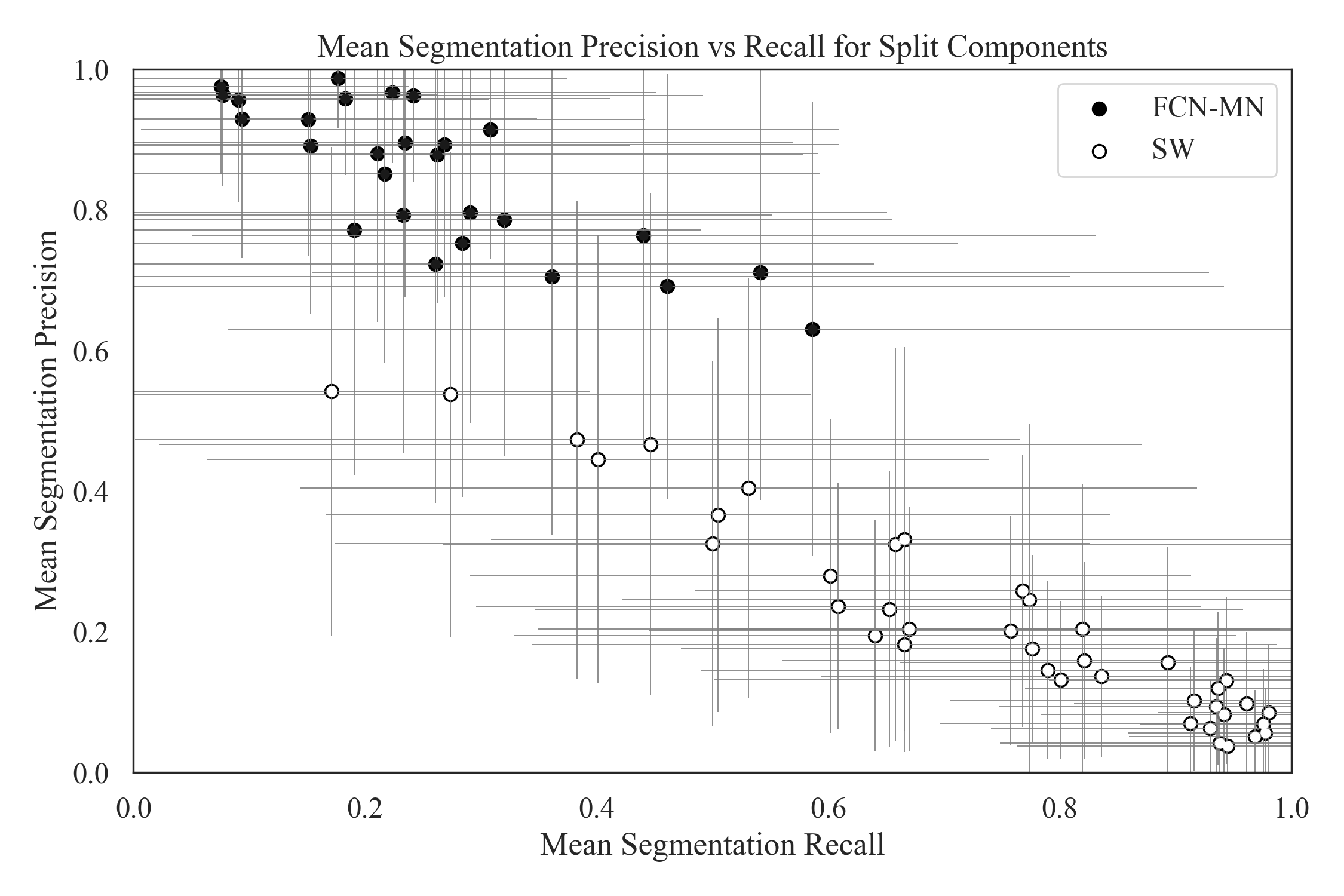}
       \caption{}
       \label{fig:Figure5-b}
  \end{subfigure}
\caption{
Segmentation Precision-Recall scatterplots reporting the results for FCN-MN and SW in black and white, respectively, with dots representing the segmentation precision and segmentation recall average over all images in the test set (and bars representing standard deviations) with one dot per hyper-parameter configuration ($27$ for FCN-MN and $40$ for SW). In (a) averages were computed over the segmentation precision and recall of all $\alpha=0.5$ true positive components, while in (b), averages were computed over the segmentation precision and recall of the $\alpha=0.5$ split components. Recall and precision standard deviations are represented by the horizontal and vertical grey error bars, respectively.
    }%
  \label{fig:Figure5}%
\end{figure}

Figures~\ref{fig:Figure5-a}~and~\ref{fig:Figure5-b} show scatter plots for segmentation-precision and segmentation-recall for the $\alpha=0.5$ \emph{true positive} and \emph{split} components in all $140$ masks of the test images, respectively. These correspond to their respective columns of (a non-summarized version of) Table~\ref{tab:Table3} with black and white dots representing the values of FCN-MN and SW detection models, respectively. The position of each dot in the plot corresponds to the mean segmentation-precision and mean segmentation-recall over all the true positive components (splitted components, respectively) of the masks produced by the detection model associated to that dot. The standard deviation of the recall (precision) is shown as a horizontal (vertical) bar.

In Figure~\ref{fig:Figure5-a} (true positives), one can observe that all black dots (FCN-MN) are clustered in the upper-right corner of the graph, enclosed by a minimum precision and recall above $70\%$, while the white dots (SW), also clustered in the upper-right corner, are enclosed in a slightly smaller minimum precision and recall of $50\%$ and $65\%$, respectively. 

In Figure~\ref{fig:Figure5-b} (splits), one can observe a rather different scenario, with FCN-MN showing split components with precisions as large as $100\%$ but small recalls spanning the range of $10\%$ to a maximum $60\%$, while SW is showing the opposite trend recalls ranging from a low $15\%$ to a maximum of $100\%$ but small precisions all below $60\%$. When read properly, these results show, again, a better performance of FCN-MN against SW, with the former resulting in small splits (low recall) but mostly within the enclosure of the true bud (large precision), while the latter resulting in components reaching beyond the enclosure of the true bud (low precision).

\begin{figure}%
    \centering
     \includegraphics[width=\textwidth]{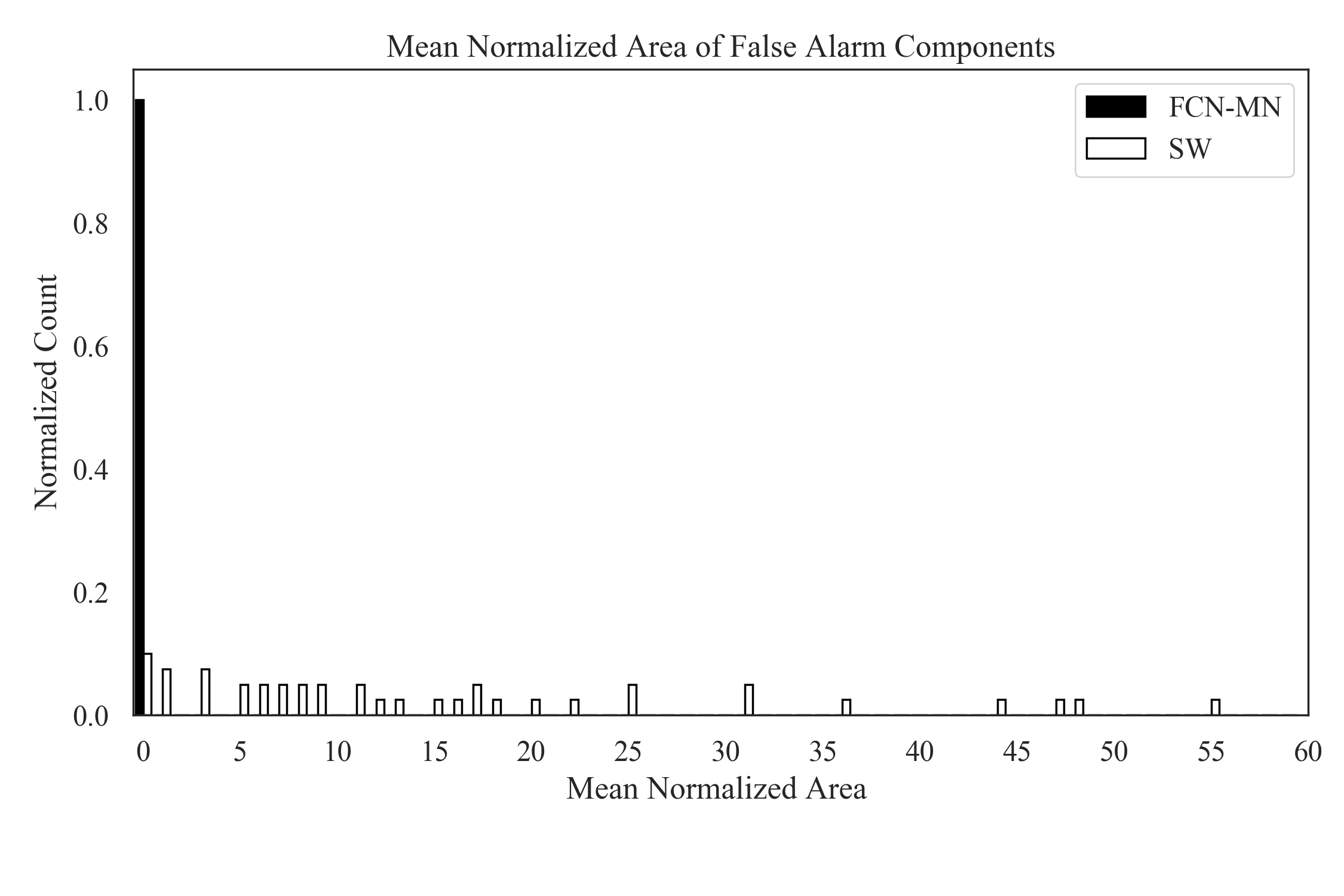}%
\caption{
FCN-MN (black bars) and SW (white bars) histograms of the mean normalized area $NA$ of false alarm components with bars representing the proportion of detection models whose mean $NA$ falls within the bin interval.
    }
\label{fig:Figure6}
\end{figure}

Figure~\ref{fig:Figure6} show a graphical representation of the segmentation results for the false alarm components, the $NA$ for each of the $27$ models of FCN-MN and each of the $40$ models of SW, i.e., for each cell in the one-before-last column of (a non-summarized version of) Table~\ref{tab:Table3}. results are grouped in two histograms, one for the FCN-MN detection models (black) and one for the SW models (white). Bars in the histogram represent the proportion of detection models whose mean $NA$ (over all false alarm components of all images) falls within the bin interval. The more concentrated to the left the better the algorithm, as this indicates that more detection models for that algorithm resulted in smaller $NA$ (on average). When compared to the histogram of SW, one can observe that the histogram for FCN-MN is considerably more concentrated towards the left, with all FCN-MN models concentrated in a single bar at the left-most interval of $[0.0, 1.0)$. For SW, the situation is rather different with bars at intervals as far to the right as $[57.0, 58.0)$, that is, detection models with areas as large as $58$ times the bud area. These high values correspond to SW models with large window sizes, e.g., 1000px, that for low thresholds are classified as bud patches, rendering all its pixels as bud pixels.

\subsubsection{Detailed analysis of localization metrics}

\begin{figure}[H]%
    \centering
 \includegraphics[width=\textwidth]{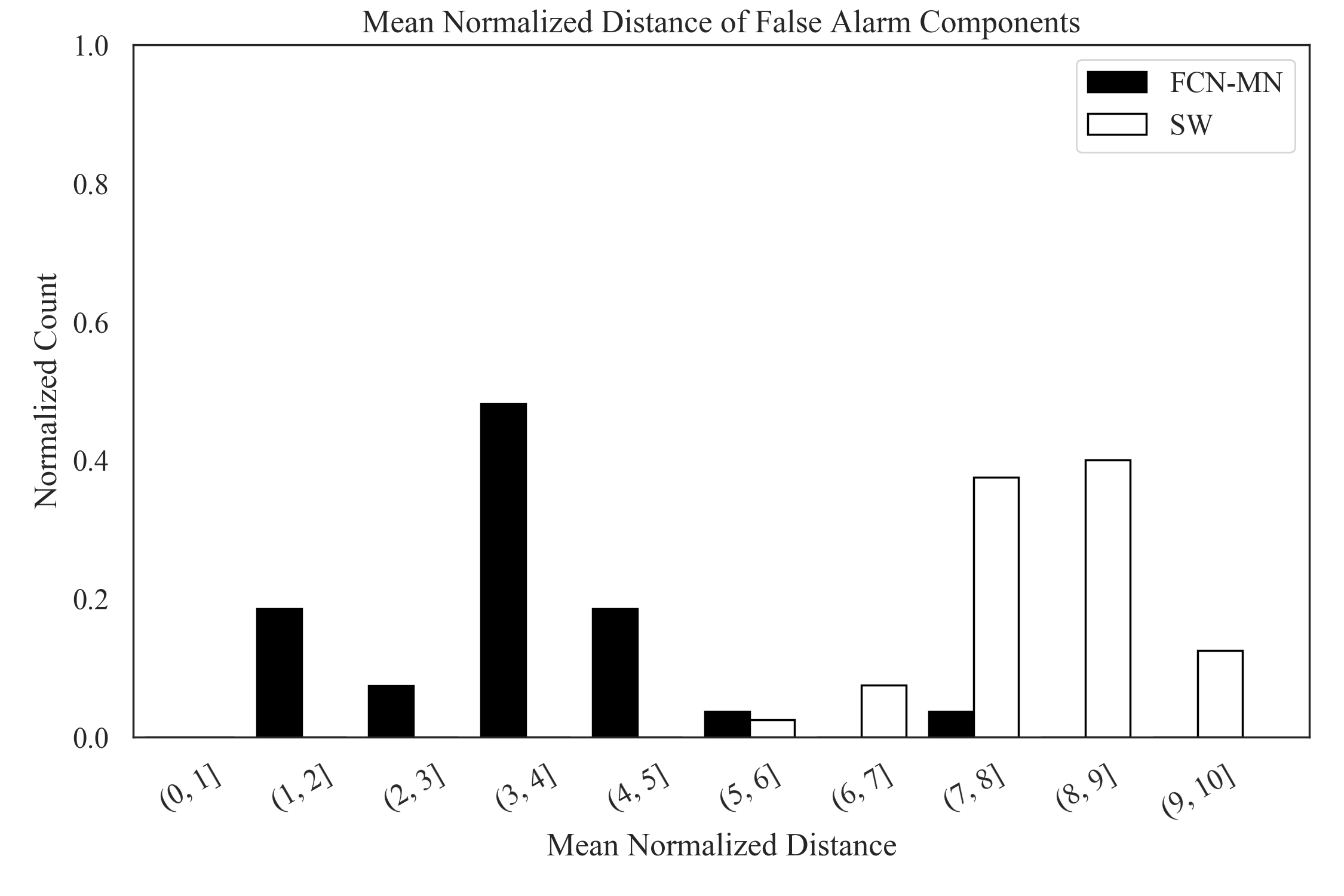}%
\caption{
FCN-MN (black bars) and SW (white bars) histograms of mean normalized distance $ND$ over all false alarm components with bars representing the proportion of detection models whose mean $ND$ falls within the bin interval.
    }
\label{fig:Figure7}
\end{figure}

To conclude, this subsection presents a graphical representation of the localization results reported in Table~\ref{tab:Table3}, that is, the \emph{normalized distance} ($ND$) only for the $\alpha=0.5$ false alarms. 
Figure~\ref{fig:Figure7} summarizes the $ND$ values reported in the corresponding column of the (non-summarized version of) Table~\ref{tab:Table3} in the form of two histograms, one for FCN-MN (black) and one for SW (white). Bars in the histogram represent the proportion of detection models ($27$ for FCN-MN and $40$ for SW) whose mean $ND$ falls within the bin interval. The more concentrated to the left the better the algorithm, as this indicates that more detection models for that algorithm resulted in smaller $ND$ (on average).
Here, again, the advantage of FCN-MN over SW is clear, with the histogram for FCN-MN more concentrated in the left-most part than that of SW, with the FCN-MN histogram running from the $(0,1]$ to the $(7,8]$ bin, and the SW histogram running from the $(5,6]$ towards the $(9,10]$ bin; and their respective maximums are at $(3,4]$ and $(7,8]$, respectively, indicating that most FCN false alarms are at a distance of $3$ to $4$ bud diameters, while most SW’s false alarms are at $7$ to $8$ bud diameters. 

\section{Discussion and Conclusions}
\label{sec:discussion}

This section discusses the results obtained by the proposed approach in the context of the problem of grapevine bud detection and its impact as a tool for measuring viticultural variables of interest. The discussion is complemented with some highlights of the most important conclusions together with some potential lines of future work.

This work introduces FCN-MN, a Fully Convolutional Network with MobileNet architecture for the detection of grapevine buds in 2D images captured in natural field conditions in winter (i.e., no leaves or bunches) containing a maximum of one bud.

The experimental results confirmed our main hypothesis: that the detection quality achieved by FCN-MN is improved over the \emph{sliding windows} detector (SW) in all three detection aspects: segmentation, correspondence identification and localization. Being SW the best bud detector known to these authors, one can conclude that FCN-MN is a strong contender in the state-of-the-art for bud detectors. 
However, improving over SW is not enough to prove the practical impact of the proposed detection algorithm, as it may still result in a bud detector with limitations for addressing the \emph{quality} requirements of practical measurements of bud-related variables (c.f. in Table~\ref{tab:Table1}). 

Quality performance could be assessed by the metrics reported in Table~\ref{tab:Table3}. In the best case, when correct detections (true positives) are considered to be those that overlap true buds with an $IoU$ of at least $0.5$, FCN-MN shows a detection-precision and detection-recall of $88.6\%$ and $95.0\%$, respectively, a mean (and standard deviation) segmentation-precision and segmentation-recall for true positives of $98.2\%(5.1)$ and $98.2\%(5.5)$, respectively; and for splits $98.8\%(7.2)$ and $58.6\%(50.5)$, respectively. For false alarms, it shows a minimum $NA$ of $0.04(0.09)$ and a minimum $ND$ of $1.10(0.65)$. 
However, each of these best cases occur for different FCN-MN detectors. A better assessment must be conducted for a single detector. A balanced choice is the detector FCN-MN$_{16s}^{0.6}$. This detector reaches detection-precision and detection-recall of $88.6\%$ and $88.6\%$, respectively, meaning than only $11.4\%$ of all the detected connected components over all test images are false positives, and that only $11.4\%$ of all true buds could not be detected (i.e., are false negatives).
As expected, for splits it resulted in small mean $IoU$ of $18.6\%(19.5)$, corresponding to a mean segmentation precision of $89.3\%(21.7)$ and a mean segmentation recall of $26.9\%(34.1)$. A small recall or small precision is expected for a small $IoU$, but a large precision and small recall is preferred as it corresponds to small components mostly within the true bud. True positives resulted in mean $IoU$ of $83.1\%(9.4)$, considerably larger than the minimum threshold of $0.5$, with correspondingly large mean segmentation precision and recall of $92.8\%(6.7)$ and recall of $89.3\%(10.2)$, respectively. 
The false alarm results for this detector showed an $NA=0.08$ and $ND=1.1$, showing that these components are rather small covering only an area that is $8\%$ in size of the total bud area (on average) and distant to the true bud by only $1.1 (0.65)$ diameters, on average.

With one select best model it is now possible to  assess the impact of its good metrics on the quality for the measurement of different bud related variables. For brevity, this point is discussed for three variables selected from Table~\ref{tab:Table1}: \emph{bud number}, \emph{bud area}, and \emph{internode length}.

The case of \emph{bud number}, for example, requires identifying correspondences for buds in the scene, so its quality will be impacted only by the metrics of detection precision and recall ($88.6\%$ and $88.6\%$, respectively). To evaluate this impact, it is considered that a plant has approximately $240$ buds on average. The number of buds per plant depends on many factors, such as training system, grape variety, type of treatment, time of year, among others, so this value is defined as indicative to achieve an approximate analysis. For this particular case, a detection-precision of $88.6\%$ would result in $27$ buds counted in excess per plant, while a recall of $88.6\%$ would result in the omission of $27$ buds in the count; resulting in a bud count that exactly matches the true count. This suggests that this particular hand-picked model may be the best for the specific task of measuring bud number, as long as these values of precision and recall remain similar. 
However, even if it is indeed the case that these good results generalize to practical situations, the approach still presents some practical limitations for the measurement of bud number. Namely, it is incapable of associating counts of the same bud appearing in two different images, limiting the scalability to massive measurements of the bud count of a plant or plot.

The second variable of interest considered is \emph{bud area}, where, in addition to identifying correspondences for the buds of a scene, it is necessary to segment it to estimate its area in pixels. Correspondence identification analysis is analogous to bud counting, so now only segmentation metrics are discussed.
A thorough error analysis requires an estimation of both \emph{false negative pixels} or $fnx$ corresponding to undetected bud pixels; and \emph{false positive pixels} or $fpx$ corresponding to non-bud detected pixels. 
These, however, are impossible to compute exactly from existing segmentation metrics. 
False negative pixels are encoded in the segmentation-recalls of \emph{true positives}  and \emph{splits}, corresponding to the sum of their complements. However, we have only their segmentation-recall means, $R_S^{TP} = 89.3\%$ and $R_S^{S} = 26.9\%$, which adds to more than $100\%$. As an approximation we assume $fnx=0$.
False positive pixels are encoded in the normalized area of \emph{false alarms} and  the segmentation-precisions of \emph{true positives} and \emph{splits}. Precision, however, is normalized by the detected area, not the true bud area. For the lack of a better solution, one can approximate $fpx$ ignoring this distinction, computing $fpx$ as the sum of $8\%$, the (mean) false alarms $NA$, with $7.2\%$ and $10.7\%$, the complements of the mean segmentation-precision of  true positives and splits ($P_S^{TP}=92.8\%$ and $P_S^{S}=89.3\%$, respectively). This results in an approximate $fpx$ equal to $25.9\%$, which given $fnx=0$, corresponds to the total error. 
For illustrative purposes, we see that this error is equivalent to the precision error resulting from measuring the area of a bud with a caliper. If we assume that the shape of a bud fits a circle, and that the typical diameter of a bud is $5$ mm, the resulting area is $19.63 mm^2$. Since a caliper has an accuracy of $0.1 mm$, the area precision error would be $\pm 1.7 mm^2$, equivalent to $8.6\%$ of the total area, to which one should add the error of manual measurement resulting from assuming a circular bud shape. From these, one can conclude that, modulo the approximations, both the FCN-MN and caliper errors are equivalent.

As in the case of counting, these good results in measurement precision are limited to achieve a practical use of this type of measurement because it is impossible to automatically associate area measurements of the same bud in two different images, making it difficult to systematically measure this variable for the buds of a plant or plot. Furthermore, in this case, the areas obtained are in pixels, which need to be converted into length or area magnitudes.

Finally, the case of \emph{internode length} is considered, estimated by the distance between buds of the same branch (by the closeness between buds and nodes), which involves the operations of correspondence identification and localization. Again, correspondence identification analysis is analogous to bud counting, which in this case will result in the reporting of more than one distance due to the detection of more than one component per bud. Among these distances, it is understood that the worst case can occur between two false alarms when they are at the farthest side to the other bud, at a distance $ND$. On average, $ND$ is $1.1$ bud diameters, equivalent to $5.5$mm after taking a typical vine bud diameter to be $5$mm, resulting in a $7.3\%$ error in estimating the distance between buds/nodes by taking the typical bud distances to be approximately $15$cm. 
An important limitation of our approach for achieving a practical use of this measurement is the possibility of determining when two buds are on the same branch, which requires knowledge of the plant structure. Furthermore, with our method, only the distance projected in the image plane could be measured, which can arbitrarily differ from the actual distance in 3D.

The greatest impact errors occur because of the excess or omission of connected components, with the excess error exacerbated by the fact of associating detected buds with individual connected components. A possible improvement to mitigate these errors would be to apply some post-processing. 
One such post-processing is \emph{spatial clustering} of connected components grouping them by proximity. One could expect this to improve the results based on the small areas of split and false alarm components. First, due to the closeness of the false alarms to the true bud (small $ND$) --as well as the splits and true positive components (overlapping with it)--, and the fact that true buds in real plants are typically tens or even hundreds of bud diameters apart, one could expect that a simple spatial clustering of the components would connect all of them together as a single, and correct, bud detection. Second, due to their small area --if clustered together-- the false alarm components would only slightly reduce segmentation precision.

Another possible post-processing would be to rule out small connected components, for example, whose area in pixels normalized to the total detected area (sum of the areas of all connected components) is less than a certain threshold. Improvements could be expected with this post-processing, since the results in this work show that false alarms present small areas in relation to the true bud. Lastly, connected component filters could be considered based on plant structure, for example, ruling out connected components that are far away from (or do not overlap with) branches.

One could also consider in future works some improvements to overcome the limitations for practical use mentioned above: (i) no associations between plant parts of different images, (ii) distance and area measurements in pixels, (iii) only 2D geometry, (iv) lack of knowledge of underlying plant structure, and (v) need of images with no leaves. 

One could also extend to buds the work of \citet{santos2020grape} that addresses limitation (i) for grape bunches. Limitation (ii) could be easily addressed by adding to the visual scene some marker with known dimensions. This, however, requires such a marker in every image captured, a problem that could be overcome by first producing a calibrated 3D reconstruction of the scene, i.e., a 3D reconstruction calibrated with a single marker in one of its frames \citep{hartley2003multiple, moons20093d}. In this way, every 2D image could be calibrated against the 3D model, omitting the need for a marker. In addition, a 3D reconstruction of the scene could address limitation (iii) by locating the detected buds in 3D space, following, for instance, the approach taken by \citet{diaz2018grapevine}. Finally, a solution to limitations (iv) and (v) would require an integrated approach involving the detection in 3D of branches and leaves, respectively. 

To end, future research could examine a comprehensive evaluation of the FCN-MN detector over images with multiple bud cases. This challenging task involves the creation of a new corpus, the training of new FCN-MN models, and the systematic evaluation of experiments. Such work may bring about a greater impact to FCN-MN, by being able to validate its performance over a broader range of practical cases that may take place in real vineyards.

\section*{Acknowledgments}

This work was funded by the Argentinean \emph{Universidad Tecnológica Nacional} (UTN), the National Council of Scientific and Technical Research (CONICET), and the National Fund for Scientific and Technological Promotion (FONCyT).

\bibliography{2020-DeepBudDetection-CEA-R2}

\begin{thebibliography}{46}
\expandafter\ifx\csname natexlab\endcsname\relax\def\natexlab#1{#1}\fi
\providecommand{\url}[1]{\texttt{#1}}
\providecommand{\href}[2]{#2}
\providecommand{\path}[1]{#1}
\providecommand{\DOIprefix}{doi:}
\providecommand{\ArXivprefix}{arXiv:}
\providecommand{\URLprefix}{URL: }
\providecommand{\Pubmedprefix}{pmid:}
\providecommand{\doi}[1]{\href{http://dx.doi.org/#1}{\path{#1}}}
\providecommand{\Pubmed}[1]{\href{pmid:#1}{\path{#1}}}
\providecommand{\bibinfo}[2]{#2}
\ifx\xfnm\relax \def\xfnm[#1]{\unskip,\space#1}\fi
\bibitem[{Berenstein et~al.(2010)Berenstein, Shahar, Shapiro and
  Edan}]{berenstein2010grape}
\bibinfo{author}{Berenstein, R.}, \bibinfo{author}{Shahar, O.B.},
  \bibinfo{author}{Shapiro, A.}, \bibinfo{author}{Edan, Y.},
  \bibinfo{year}{2010}.
\newblock \bibinfo{title}{Grape clusters and foliage detection algorithms for
  autonomous selective vineyard sprayer}.
\newblock \bibinfo{journal}{Intelligent Service Robotics} \bibinfo{volume}{3},
  \bibinfo{pages}{233--243}.
\bibitem[{Bramley(2009)}]{bramley2009lessons}
\bibinfo{author}{Bramley, R.G.}, \bibinfo{year}{2009}.
\newblock \bibinfo{title}{Lessons from nearly 20 years of precision agriculture
  research, development, and adoption as a guide to its appropriate
  application}.
\newblock \bibinfo{journal}{Crop and Pasture Science} \bibinfo{volume}{60},
  \bibinfo{pages}{197--217}.
\bibitem[{Collins et~al.(2020)Collins, Wang, Lesefko, De~Bei and
  Fuentes}]{collins2020effects}
\bibinfo{author}{Collins, C.}, \bibinfo{author}{Wang, X.},
  \bibinfo{author}{Lesefko, S.}, \bibinfo{author}{De~Bei, R.},
  \bibinfo{author}{Fuentes, S.}, \bibinfo{year}{2020}.
\newblock \bibinfo{title}{Effects of canopy management practices on grapevine
  bud fruitfulness}.
\newblock \bibinfo{journal}{OENO One} \bibinfo{volume}{54},
  \bibinfo{pages}{313--325}.
\bibitem[{Diago et~al.(2012)Diago, Correa, Mill{\'a}n, Barreiro, Valero and
  Tardaguila}]{diago2012grapevine}
\bibinfo{author}{Diago, M.P.}, \bibinfo{author}{Correa, C.},
  \bibinfo{author}{Mill{\'a}n, B.}, \bibinfo{author}{Barreiro, P.},
  \bibinfo{author}{Valero, C.}, \bibinfo{author}{Tardaguila, J.},
  \bibinfo{year}{2012}.
\newblock \bibinfo{title}{Grapevine yield and leaf area estimation using
  supervised classification methodology on rgb images taken under field
  conditions}.
\newblock \bibinfo{journal}{Sensors} \bibinfo{volume}{12},
  \bibinfo{pages}{16988--17006}.
\bibitem[{D{\'\i}az et~al.(2018)D{\'\i}az, P{\'e}rez, Miatello and
  Bromberg}]{diaz2018grapevine}
\bibinfo{author}{D{\'\i}az, C.A.}, \bibinfo{author}{P{\'e}rez, D.S.},
  \bibinfo{author}{Miatello, H.}, \bibinfo{author}{Bromberg, F.},
  \bibinfo{year}{2018}.
\newblock \bibinfo{title}{Grapevine buds detection and localization in 3d space
  based on structure from motion and 2d image classification}.
\newblock \bibinfo{journal}{Computers in Industry} \bibinfo{volume}{99},
  \bibinfo{pages}{303--312}.
\bibitem[{Garcia-Garcia et~al.(2018)Garcia-Garcia, Orts-Escolano, Oprea,
  Villena-Martinez, Martinez-Gonzalez and Garcia-Rodriguez}]{garcia2018survey}
\bibinfo{author}{Garcia-Garcia, A.}, \bibinfo{author}{Orts-Escolano, S.},
  \bibinfo{author}{Oprea, S.}, \bibinfo{author}{Villena-Martinez, V.},
  \bibinfo{author}{Martinez-Gonzalez, P.}, \bibinfo{author}{Garcia-Rodriguez,
  J.}, \bibinfo{year}{2018}.
\newblock \bibinfo{title}{A survey on deep learning techniques for image and
  video semantic segmentation}.
\newblock \bibinfo{journal}{Applied Soft Computing} \bibinfo{volume}{70},
  \bibinfo{pages}{41--65}.
\bibitem[{Grimm et~al.(2019)Grimm, Herzog, Rist, Kicherer, T{\"o}pfer and
  Steinhage}]{grimm2019adaptable}
\bibinfo{author}{Grimm, J.}, \bibinfo{author}{Herzog, K.},
  \bibinfo{author}{Rist, F.}, \bibinfo{author}{Kicherer, A.},
  \bibinfo{author}{T{\"o}pfer, R.}, \bibinfo{author}{Steinhage, V.},
  \bibinfo{year}{2019}.
\newblock \bibinfo{title}{An adaptable approach to automated visual detection
  of plant organs with applications in grapevine breeding}.
\newblock \bibinfo{journal}{Biosystems Engineering} \bibinfo{volume}{183},
  \bibinfo{pages}{170--183}.
\bibitem[{Han(2013)}]{han2013comparison}
\bibinfo{author}{Han, D.}, \bibinfo{year}{2013}.
\newblock \bibinfo{title}{Comparison of commonly used image interpolation
  methods}, in: \bibinfo{booktitle}{Proceedings of the 2nd international
  conference on computer science and electronics engineering},
  \bibinfo{organization}{Atlantis Press}.
\bibitem[{Hartley and Zisserman(2003)}]{hartley2003multiple}
\bibinfo{author}{Hartley, R.}, \bibinfo{author}{Zisserman, A.},
  \bibinfo{year}{2003}.
\newblock \bibinfo{title}{Multiple view geometry in computer vision}.
\newblock \bibinfo{publisher}{Cambridge university press}.
\bibitem[{Herzog et~al.(2014a)Herzog, Kicherer and
  T{\"o}pfer}]{herzog2014objective}
\bibinfo{author}{Herzog, K.}, \bibinfo{author}{Kicherer, A.},
  \bibinfo{author}{T{\"o}pfer, R.}, \bibinfo{year}{2014}a.
\newblock \bibinfo{title}{Objective phenotyping the time of bud burst by
  analyzing grapevine field images}, in: \bibinfo{booktitle}{XI International
  Conference on Grapevine Breeding and Genetics 1082}, pp.
  \bibinfo{pages}{379--385}.
\bibitem[{Herzog et~al.(2014b)}]{herzog2014initial}
\bibinfo{author}{Herzog, K.}, et~al., \bibinfo{year}{2014}b.
\newblock \bibinfo{title}{Initial steps for high-throughput phenotyping in
  vineyards}.
\newblock \bibinfo{journal}{Australian and New Zealand Grapegrower and
  Winemaker} , \bibinfo{pages}{54}.
\bibitem[{Hirano et~al.(2006)Hirano, Garcia, Sukthankar and
  Hoogs}]{hirano2006industry}
\bibinfo{author}{Hirano, Y.}, \bibinfo{author}{Garcia, C.},
  \bibinfo{author}{Sukthankar, R.}, \bibinfo{author}{Hoogs, A.},
  \bibinfo{year}{2006}.
\newblock \bibinfo{title}{Industry and object recognition: Applications,
  applied research and challenges}, in: \bibinfo{booktitle}{Toward
  category-level object recognition}. \bibinfo{publisher}{Springer}, pp.
  \bibinfo{pages}{49--64}.
\bibitem[{Howard et~al.(2017)Howard, Zhu, Chen, Kalenichenko, Wang, Weyand,
  Andreetto and Adam}]{howard2017mobilenets}
\bibinfo{author}{Howard, A.G.}, \bibinfo{author}{Zhu, M.},
  \bibinfo{author}{Chen, B.}, \bibinfo{author}{Kalenichenko, D.},
  \bibinfo{author}{Wang, W.}, \bibinfo{author}{Weyand, T.},
  \bibinfo{author}{Andreetto, M.}, \bibinfo{author}{Adam, H.},
  \bibinfo{year}{2017}.
\newblock \bibinfo{title}{Mobilenets: Efficient convolutional neural networks
  for mobile vision applications}.
\newblock \bibinfo{journal}{arXiv preprint arXiv:1704.04861} .
\bibitem[{Jaccard(1912)}]{jaccard1912distribution}
\bibinfo{author}{Jaccard, P.}, \bibinfo{year}{1912}.
\newblock \bibinfo{title}{The distribution of the flora in the alpine zone. 1}.
\newblock \bibinfo{journal}{New phytologist} \bibinfo{volume}{11},
  \bibinfo{pages}{37--50}.
\bibitem[{Kahng et~al.(2017)Kahng, Andrews, Kalro and Chau}]{kahng2017cti}
\bibinfo{author}{Kahng, M.}, \bibinfo{author}{Andrews, P.Y.},
  \bibinfo{author}{Kalro, A.}, \bibinfo{author}{Chau, D.H.P.},
  \bibinfo{year}{2017}.
\newblock \bibinfo{title}{A cti v is: Visual exploration of industry-scale deep
  neural network models}.
\newblock \bibinfo{journal}{IEEE transactions on visualization and computer
  graphics} \bibinfo{volume}{24}, \bibinfo{pages}{88--97}.
\bibitem[{Kaymak and U{\c{c}}ar(2019)}]{kaymak2019brief}
\bibinfo{author}{Kaymak, {\c{C}}.}, \bibinfo{author}{U{\c{c}}ar, A.},
  \bibinfo{year}{2019}.
\newblock \bibinfo{title}{A brief survey and an application of semantic image
  segmentation for autonomous driving}, in: \bibinfo{booktitle}{Handbook of
  Deep Learning Applications}. \bibinfo{publisher}{Springer}, pp.
  \bibinfo{pages}{161--200}.
\bibitem[{Kornblith et~al.(2019)Kornblith, Shlens and Le}]{kornblith2019better}
\bibinfo{author}{Kornblith, S.}, \bibinfo{author}{Shlens, J.},
  \bibinfo{author}{Le, Q.V.}, \bibinfo{year}{2019}.
\newblock \bibinfo{title}{Do better imagenet models transfer better?}, in:
  \bibinfo{booktitle}{Proceedings of the IEEE conference on computer vision and
  pattern recognition}, pp. \bibinfo{pages}{2661--2671}.
\bibitem[{Lampert et~al.(2008)Lampert, Blaschko and
  Hofmann}]{lampert2008beyond}
\bibinfo{author}{Lampert, C.H.}, \bibinfo{author}{Blaschko, M.B.},
  \bibinfo{author}{Hofmann, T.}, \bibinfo{year}{2008}.
\newblock \bibinfo{title}{Beyond sliding windows: Object localization by
  efficient subwindow search}, in: \bibinfo{booktitle}{2008 IEEE conference on
  computer vision and pattern recognition}, \bibinfo{organization}{IEEE}. pp.
  \bibinfo{pages}{1--8}.
\bibitem[{Litjens et~al.(2017)Litjens, Kooi, Bejnordi, Setio, Ciompi,
  Ghafoorian, Van Der~Laak, Van~Ginneken and S{\'a}nchez}]{litjens2017survey}
\bibinfo{author}{Litjens, G.}, \bibinfo{author}{Kooi, T.},
  \bibinfo{author}{Bejnordi, B.E.}, \bibinfo{author}{Setio, A.A.A.},
  \bibinfo{author}{Ciompi, F.}, \bibinfo{author}{Ghafoorian, M.},
  \bibinfo{author}{Van Der~Laak, J.A.}, \bibinfo{author}{Van~Ginneken, B.},
  \bibinfo{author}{S{\'a}nchez, C.I.}, \bibinfo{year}{2017}.
\newblock \bibinfo{title}{A survey on deep learning in medical image analysis}.
\newblock \bibinfo{journal}{Medical image analysis} \bibinfo{volume}{42},
  \bibinfo{pages}{60--88}.
\bibitem[{Long et~al.(2015)Long, Shelhamer and Darrell}]{long2015fully}
\bibinfo{author}{Long, J.}, \bibinfo{author}{Shelhamer, E.},
  \bibinfo{author}{Darrell, T.}, \bibinfo{year}{2015}.
\newblock \bibinfo{title}{Fully convolutional networks for semantic
  segmentation}, in: \bibinfo{booktitle}{Proceedings of the IEEE conference on
  computer vision and pattern recognition}, pp. \bibinfo{pages}{3431--3440}.
\bibitem[{Matese and Di~Gennaro(2015)}]{matese2015technology}
\bibinfo{author}{Matese, A.}, \bibinfo{author}{Di~Gennaro, S.F.},
  \bibinfo{year}{2015}.
\newblock \bibinfo{title}{Technology in precision viticulture: A state of the
  art review}.
\newblock \bibinfo{journal}{International journal of wine research}
  \bibinfo{volume}{7}, \bibinfo{pages}{69--81}.
\bibitem[{May(2000)}]{may2000bud}
\bibinfo{author}{May, P.}, \bibinfo{year}{2000}.
\newblock \bibinfo{title}{From bud to berry, with special reference to
  inflorescence and bunch morphology in vitis vinifera l.}
\newblock \bibinfo{journal}{Australian Journal of Grape and Wine Research}
  \bibinfo{volume}{6}, \bibinfo{pages}{82--98}.
\bibitem[{Moons et~al.(2009)Moons, Van~Gool and Vergauwen}]{moons20093d}
\bibinfo{author}{Moons, T.}, \bibinfo{author}{Van~Gool, L.},
  \bibinfo{author}{Vergauwen, M.}, \bibinfo{year}{2009}.
\newblock \bibinfo{title}{3D Reconstruction from Multiple Images: Principles}.
\newblock \bibinfo{publisher}{Now Publishers Inc}.
\bibitem[{Ning et~al.(2017)Ning, Zhou, Song and Tang}]{ning2017inception}
\bibinfo{author}{Ning, C.}, \bibinfo{author}{Zhou, H.}, \bibinfo{author}{Song,
  Y.}, \bibinfo{author}{Tang, J.}, \bibinfo{year}{2017}.
\newblock \bibinfo{title}{Inception single shot multibox detector for object
  detection}, in: \bibinfo{booktitle}{2017 IEEE International Conference on
  Multimedia \& Expo Workshops (ICMEW)}, \bibinfo{organization}{IEEE}. pp.
  \bibinfo{pages}{549--554}.
\bibitem[{Noyce et~al.(2016)Noyce, Steel, Harper and Wood}]{noyce2016basis}
\bibinfo{author}{Noyce, P.W.}, \bibinfo{author}{Steel, C.C.},
  \bibinfo{author}{Harper, J.D.}, \bibinfo{author}{Wood, R.M.},
  \bibinfo{year}{2016}.
\newblock \bibinfo{title}{The basis of defoliation effects on reproductive
  parameters in vitis vinifera l. cv. chardonnay lies in the latent bud}.
\newblock \bibinfo{journal}{American Journal of Enology and Viticulture}
  \bibinfo{volume}{67}, \bibinfo{pages}{199--205}.
\bibitem[{Nuske et~al.(2011)Nuske, Achar, Bates, Narasimhan and
  Singh}]{nuske2011yield}
\bibinfo{author}{Nuske, S.}, \bibinfo{author}{Achar, S.},
  \bibinfo{author}{Bates, T.}, \bibinfo{author}{Narasimhan, S.},
  \bibinfo{author}{Singh, S.}, \bibinfo{year}{2011}.
\newblock \bibinfo{title}{Yield estimation in vineyards by visual grape
  detection}, in: \bibinfo{booktitle}{2011 IEEE/RSJ International Conference on
  Intelligent Robots and Systems}, \bibinfo{organization}{IEEE}. pp.
  \bibinfo{pages}{2352--2358}.
\bibitem[{Pan and Yang(2009)}]{pan2009survey}
\bibinfo{author}{Pan, S.J.}, \bibinfo{author}{Yang, Q.}, \bibinfo{year}{2009}.
\newblock \bibinfo{title}{A survey on transfer learning}.
\newblock \bibinfo{journal}{IEEE Transactions on knowledge and data
  engineering} \bibinfo{volume}{22}, \bibinfo{pages}{1345--1359}.
\bibitem[{P{\'e}rez et~al.(2017)P{\'e}rez, Bromberg and Diaz}]{perez2017image}
\bibinfo{author}{P{\'e}rez, D.S.}, \bibinfo{author}{Bromberg, F.},
  \bibinfo{author}{Diaz, C.A.}, \bibinfo{year}{2017}.
\newblock \bibinfo{title}{Image classification for detection of winter
  grapevine buds in natural conditions using scale-invariant features
  transform, bag of features and support vector machines}.
\newblock \bibinfo{journal}{Computers and electronics in agriculture}
  \bibinfo{volume}{135}, \bibinfo{pages}{81--95}.
\bibitem[{Rudolph et~al.(2018)Rudolph, Herzog, T{\"o}pfer and
  Steinhage}]{rudolph2018efficient}
\bibinfo{author}{Rudolph, R.}, \bibinfo{author}{Herzog, K.},
  \bibinfo{author}{T{\"o}pfer, R.}, \bibinfo{author}{Steinhage, V.},
  \bibinfo{year}{2018}.
\newblock \bibinfo{title}{Efficient identification, localization and
  quantification of grapevine inflorescences in unprepared field images using
  fully convolutional networks}.
\newblock \bibinfo{journal}{arXiv preprint arXiv:1807.03770} .
\bibitem[{S{\'a}nchez and Dokoozlian(2005)}]{sanchez2005bud}
\bibinfo{author}{S{\'a}nchez, L.A.}, \bibinfo{author}{Dokoozlian, N.K.},
  \bibinfo{year}{2005}.
\newblock \bibinfo{title}{Bud microclimate and fruitfulness in vitis vinifera
  l.}
\newblock \bibinfo{journal}{American Journal of Enology and Viticulture}
  \bibinfo{volume}{56}, \bibinfo{pages}{319--329}.
\bibitem[{Santos et~al.(2020)Santos, de~Souza, dos Santos and
  Avila}]{santos2020grape}
\bibinfo{author}{Santos, T.T.}, \bibinfo{author}{de~Souza, L.L.},
  \bibinfo{author}{dos Santos, A.A.}, \bibinfo{author}{Avila, S.},
  \bibinfo{year}{2020}.
\newblock \bibinfo{title}{Grape detection, segmentation, and tracking using
  deep neural networks and three-dimensional association}.
\newblock \bibinfo{journal}{Computers and Electronics in Agriculture}
  \bibinfo{volume}{170}, \bibinfo{pages}{105247}.
\bibitem[{Seng et~al.(2018)Seng, Ang, Schmidtke and Rogiers}]{seng2018computer}
\bibinfo{author}{Seng, K.P.}, \bibinfo{author}{Ang, L.M.},
  \bibinfo{author}{Schmidtke, L.M.}, \bibinfo{author}{Rogiers, S.Y.},
  \bibinfo{year}{2018}.
\newblock \bibinfo{title}{Computer vision and machine learning for viticulture
  technology}.
\newblock \bibinfo{journal}{IEEE Access} \bibinfo{volume}{6},
  \bibinfo{pages}{67494--67510}.
\bibitem[{Shelhamer et~al.(2017)Shelhamer, Long and
  Darrell}]{shelhamer2017fully}
\bibinfo{author}{Shelhamer, E.}, \bibinfo{author}{Long, J.},
  \bibinfo{author}{Darrell, T.}, \bibinfo{year}{2017}.
\newblock \bibinfo{title}{Fully convolutional networks for semantic
  segmentation}.
\newblock \bibinfo{journal}{IEEE transactions on pattern analysis and machine
  intelligence} \bibinfo{volume}{39}, \bibinfo{pages}{640--651}.
\bibitem[{Shorten and Khoshgoftaar(2019)}]{shorten2019survey}
\bibinfo{author}{Shorten, C.}, \bibinfo{author}{Khoshgoftaar, T.M.},
  \bibinfo{year}{2019}.
\newblock \bibinfo{title}{A survey on image data augmentation for deep
  learning}.
\newblock \bibinfo{journal}{Journal of Big Data} \bibinfo{volume}{6},
  \bibinfo{pages}{60}.
\bibitem[{Siam et~al.(2018)Siam, Gamal, Abdel-Razek, Yogamani and
  Jagersand}]{siam2018rtseg}
\bibinfo{author}{Siam, M.}, \bibinfo{author}{Gamal, M.},
  \bibinfo{author}{Abdel-Razek, M.}, \bibinfo{author}{Yogamani, S.},
  \bibinfo{author}{Jagersand, M.}, \bibinfo{year}{2018}.
\newblock \bibinfo{title}{Rtseg: Real-time semantic segmentation comparative
  study}, in: \bibinfo{booktitle}{2018 25th IEEE International Conference on
  Image Processing (ICIP)}, \bibinfo{organization}{IEEE}. pp.
  \bibinfo{pages}{1603--1607}.
\bibitem[{Simonyan and Zisserman(2015)}]{Simonyan2015VeryDC}
\bibinfo{author}{Simonyan, K.}, \bibinfo{author}{Zisserman, A.},
  \bibinfo{year}{2015}.
\newblock \bibinfo{title}{Very deep convolutional networks for large-scale
  image recognition}.
\newblock \bibinfo{journal}{CoRR} \bibinfo{volume}{abs/1409.1556}.
\bibitem[{Tardaguila et~al.(2012)Tardaguila, Diago, Blasco, Mill{\'a}n, Cubero,
  Garc{\'\i}a-Navarrete and Aleixos}]{tardaguila2012automatic}
\bibinfo{author}{Tardaguila, J.}, \bibinfo{author}{Diago, M.},
  \bibinfo{author}{Blasco, J.}, \bibinfo{author}{Mill{\'a}n, B.},
  \bibinfo{author}{Cubero, S.}, \bibinfo{author}{Garc{\'\i}a-Navarrete, O.},
  \bibinfo{author}{Aleixos, N.}, \bibinfo{year}{2012}.
\newblock \bibinfo{title}{Automatic estimation of the size and weight of
  grapevine berries by image analysis}, in: \bibinfo{booktitle}{Proc. CIGR
  AgEng}.
\bibitem[{Tard{\'a}guila et~al.(2012)Tard{\'a}guila, Diago, Millan, Blasco,
  Cubero and Aleixos}]{tardaguila2012applications}
\bibinfo{author}{Tard{\'a}guila, J.}, \bibinfo{author}{Diago, M.P.},
  \bibinfo{author}{Millan, B.}, \bibinfo{author}{Blasco, J.},
  \bibinfo{author}{Cubero, S.}, \bibinfo{author}{Aleixos, N.},
  \bibinfo{year}{2012}.
\newblock \bibinfo{title}{Applications of computer vision techniques in
  viticulture to assess canopy features, cluster morphology and berry size},
  in: \bibinfo{booktitle}{I International Workshop on Vineyard Mechanization
  and Grape and Wine Quality 978}, pp. \bibinfo{pages}{77--84}.
\bibitem[{Tarry et~al.(2014)Tarry, Wspanialy, Veres and
  Moussa}]{tarry2014integrated}
\bibinfo{author}{Tarry, C.}, \bibinfo{author}{Wspanialy, P.},
  \bibinfo{author}{Veres, M.}, \bibinfo{author}{Moussa, M.},
  \bibinfo{year}{2014}.
\newblock \bibinfo{title}{An integrated bud detection and localization system
  for application in greenhouse automation}, in: \bibinfo{booktitle}{2014
  Canadian Conference on Computer and Robot Vision},
  \bibinfo{organization}{IEEE}. pp. \bibinfo{pages}{344--348}.
\bibitem[{{The Australian Wine Research Institute}(a)}]{awriNDmanual1}
\bibinfo{author}{{The Australian Wine Research Institute}}, a.
\newblock \bibinfo{title}{Viticare on Farm Trials - Manual 3.1: Measuring Fruit
  Quality}. \bibinfo{edition}{1} ed.
\newblock \bibinfo{organization}{The Australian Wine Research Institute}.
\newblock \bibinfo{note}{Accessed August 2020}.
\bibitem[{{The Australian Wine Research Institute}(b)}]{awriNDmanual3}
\bibinfo{author}{{The Australian Wine Research Institute}}, b.
\newblock \bibinfo{title}{Viticare on Farm Trials - Manual 3.3: Vine Health}.
  \bibinfo{edition}{1} ed.
\newblock \bibinfo{organization}{The Australian Wine Research Institute}.
\newblock \bibinfo{note}{Accessed August 2020}.
\bibitem[{Tilgner et~al.(2019)Tilgner, Wagner, Kalischewski, Velten and
  Kummert}]{tilgner2019multi}
\bibinfo{author}{Tilgner, S.}, \bibinfo{author}{Wagner, D.},
  \bibinfo{author}{Kalischewski, K.}, \bibinfo{author}{Velten, J.},
  \bibinfo{author}{Kummert, A.}, \bibinfo{year}{2019}.
\newblock \bibinfo{title}{Multi-view fusion neural network with application in
  the manufacturing industry}, in: \bibinfo{booktitle}{2019 IEEE International
  Symposium on Circuits and Systems (ISCAS)}, \bibinfo{organization}{IEEE}. pp.
  \bibinfo{pages}{1--5}.
\bibitem[{Whalley and Shanmuganathan(2013)}]{whalley2013applications}
\bibinfo{author}{Whalley, J.}, \bibinfo{author}{Shanmuganathan, S.},
  \bibinfo{year}{2013}.
\newblock \bibinfo{title}{Applications of image processing in viticulture: A
  review} .
\bibitem[{Whelan et~al.(1996)Whelan, McBratney and
  Viscarra~Rossel}]{whelan1996spatial}
\bibinfo{author}{Whelan, B.}, \bibinfo{author}{McBratney, A.},
  \bibinfo{author}{Viscarra~Rossel, R.}, \bibinfo{year}{1996}.
\newblock \bibinfo{title}{Spatial prediction for precision agriculture}, in:
  \bibinfo{booktitle}{Proceedings of the Third International Conference on
  Precision Agriculture}, \bibinfo{organization}{Wiley Online Library}. pp.
  \bibinfo{pages}{331--342}.
\bibitem[{Xu et~al.(2014)Xu, Xun, Jia and Yang}]{xu2014detection}
\bibinfo{author}{Xu, S.}, \bibinfo{author}{Xun, Y.}, \bibinfo{author}{Jia, T.},
  \bibinfo{author}{Yang, Q.}, \bibinfo{year}{2014}.
\newblock \bibinfo{title}{Detection method for the buds on winter vines based
  on computer vision}, in: \bibinfo{booktitle}{2014 Seventh International
  Symposium on Computational Intelligence and Design},
  \bibinfo{organization}{IEEE}. pp. \bibinfo{pages}{44--48}.
\bibitem[{Zhao et~al.(2018)Zhao, Rong, Liping and Chenlong}]{zhao2018research}
\bibinfo{author}{Zhao, F.}, \bibinfo{author}{Rong, D.},
  \bibinfo{author}{Liping, L.}, \bibinfo{author}{Chenlong, L.},
  \bibinfo{year}{2018}.
\newblock \bibinfo{title}{Research on stalk crops internodes and buds
  identification based on computer vision}.
\newblock \bibinfo{journal}{MS\&E} \bibinfo{volume}{439},
  \bibinfo{pages}{032080}.

\end{thebibliography}

\end{document}